\pdfoutput=1

\documentclass[11pt]{article}

\usepackage{ACL2023}

\usepackage{times}
\usepackage{latexsym}

\usepackage[T1]{fontenc}

\usepackage[utf8]{inputenc}

\usepackage{microtype}

\usepackage{inconsolata}

\usepackage{booktabs}
\usepackage{multirow}
\usepackage{graphicx}
\usepackage{subcaption}
\usepackage{mdframed}

\newcommand{\fone}{F$_1$}
\newcommand{\boldfone}{\textbf{F}$_\mathbf{1}$}

\newcommand{\oqa}{open-domain QA}{}
\newcommand{\Oqa}{Open-domain QA}{}

\newcommand{\NQopen}{NQ-\textsc{open}}{}
\newcommand{\NaturalQopen}{Natural Questions-\textsc{open}}{}

{}

\title{Evaluating Open-Domain Question Answering\\in the Era of Large Language Models}

\author{Ehsan Kamalloo$^{\:\diamondsuit\:\clubsuit}$ \quad 
Nouha Dziri$^{\:\spadesuit}$ \quad
Charles L. A. Clarke$^{\:\clubsuit}$ \quad
Davood Rafiei$^{\:\diamondsuit}$ \\[1ex]
  $^{\diamondsuit}$ University of Alberta \quad $^{\clubsuit}$ University of Waterloo \\
  $^{\spadesuit}$ Allen Institute for Artificial Intelligence \\[1ex]
  \texttt{ekamalloo@uwaterloo.ca} \\}

\begin{document}
\maketitle
\begin{abstract}
Lexical matching remains the {\em de facto} evaluation method for open-domain question answering (QA). Unfortunately, lexical matching fails completely when a plausible candidate answer does not appear in the list of gold answers, which is increasingly the case as we shift from extractive to generative models.
The recent success of large language models (LLMs) for QA aggravates lexical matching failures since candidate answers become longer, thereby making matching with the gold answers even more challenging. Without accurate evaluation, the true progress in open-domain QA remains unknown. In this paper, we conduct a thorough analysis of various open-domain QA models, including LLMs, by manually evaluating their answers on a subset of {\NQopen}, a popular benchmark. Our assessments reveal that while the true performance of all models is significantly underestimated, the performance of the InstructGPT (zero-shot) LLM increases by nearly $+$60\%, making it on par with existing top models, and the InstructGPT (few-shot) model actually achieves a new state-of-the-art on {\NQopen}. We also find that more than 50\% of lexical matching failures are attributed to semantically equivalent answers. We further demonstrate that regex matching ranks QA models consistent with human judgments, although still suffering from unnecessary strictness. Finally, we demonstrate that automated evaluation models are a reasonable surrogate for lexical matching in some circumstances, but not for long-form answers generated by LLMs. The automated models struggle in detecting hallucinations in LLM answers and are thus unable to evaluate LLMs. At this time, there appears to be no substitute for human evaluation.\footnote{Code and data are released at \url{https://github.com/ehsk/OpenQA-eval}.}
\end{abstract}

\section{Introduction}

\begin{figure}[t]
  \centering
  \includegraphics[width=\linewidth]{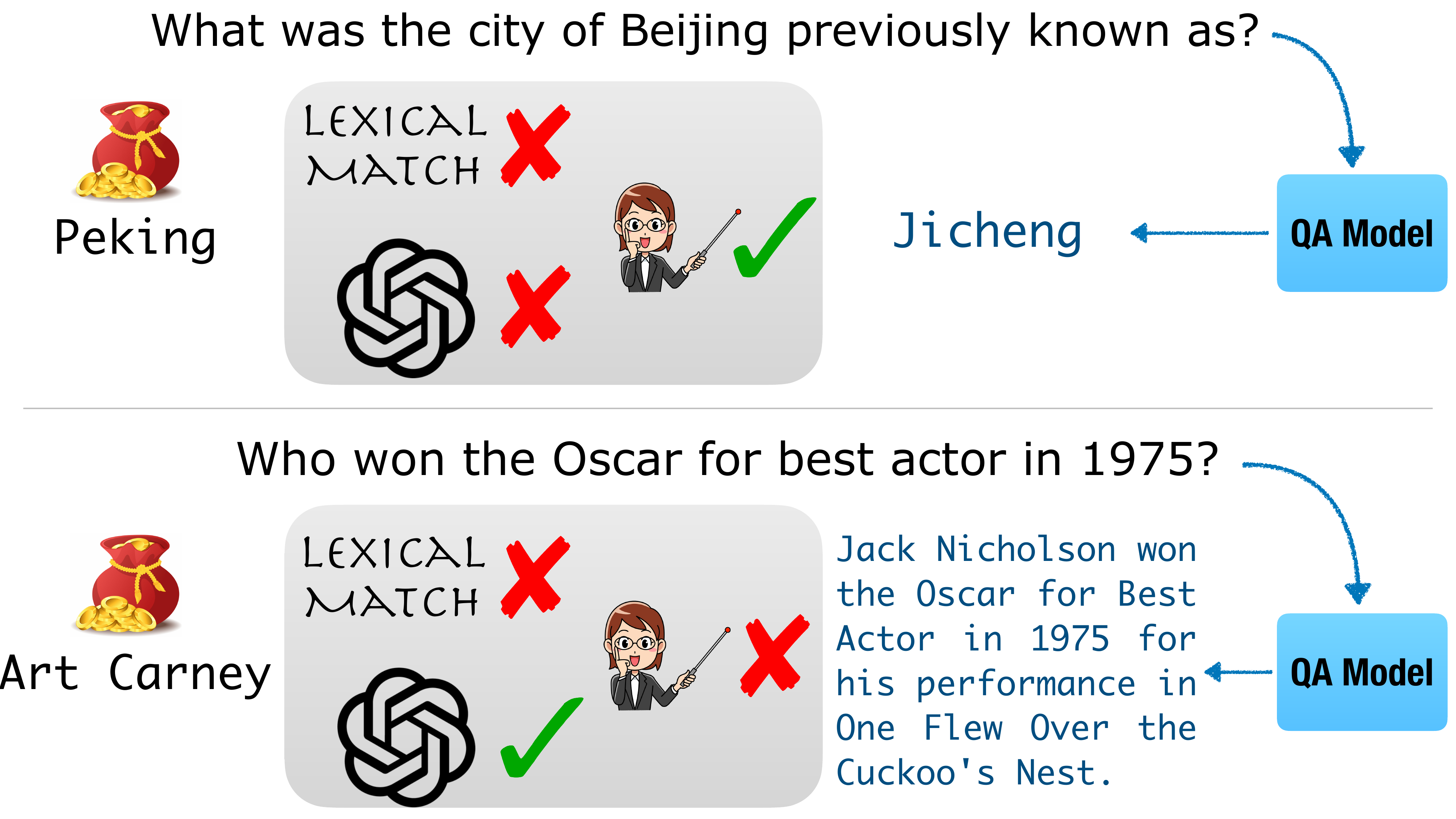}
  \caption{Examples of failures in {\oqa} evaluation. \textbf{Top:} {\em Jicheng} is a credible answer although not present in the list of gold answers. 
  Existing automated evaluation mechanisms fail to identify it as correct.
  \textbf{Bottom:} A seemingly correct but {\em unattributable} answer from InstructGPT \cite{instructgpt} for which automatic evaluation goes astray.}
  \label{fig:examples}
\end{figure}

Reliable benchmarks have been a bedrock to measuring progress in {\oqa}, the task of answering information-seeking questions over a massive text corpus. In recent years, we have seen great strides in {\oqa} by novel models (\citealt{chen-etal-2017-reading, wang2018r, clark-gardner-2018-simple, lee-etal-2019-latent, asai2020Learning, izacard-grave-2021-leveraging, izacard2021distilling, khattab-etal-2021-relevance, singh2021end, asai-etal-2022-evidentiality}; {\em inter alia}) that continue to raise state-of-the-art on well-established benchmarks such as {\NaturalQopen} \cite{lee-etal-2019-latent}. The standard procedure for evaluating {\oqa} models, borrowed from reading comprehension \cite{rajpurkar-etal-2016-squad}, is to perform lexical matching between gold answers provided in the benchmark and models' predictions. However, 
as the performance of {\oqa} approaches that of humans,\footnote{typically equipped with a search engine} these classic evaluation methods begin to fail. Such failures largely stem from the incomplete list of gold answers that do not fully cover all plausible answers. For example, in Figure~\ref{fig:examples}, ``{\em Jicheng}'' is a correct answer to {\em what was the city of Beijing previously known as?} while not annotated as a gold answer in {\NaturalQopen} ({\NQopen}; \citealt{lee-etal-2019-latent}).

With the recent success of generative QA systems in the open-domain setting \cite{izacard-grave-2021-leveraging,roberts-etal-2020-much}, 
it becomes harder for lexical matching to recognize correct answers, and in turn for us, to recognize performance differences between models. The problem is exacerbated by a tendency of Large Language Models(LLM)-based systems (\citealt{gpt3,palm,opt,black-etal-2022-gpt}; {\em inter alia}) to occasionally hallucinate plausible but incorrect answers \cite{dziri-etal-2022-origin,ye2022the}. For instance, in Figure~\ref{fig:examples}, InstructGPT \cite{instructgpt} generates ``{\em Jack Nicholson}'' in great details to answer {\em who won the oscar for best actor in 1975?} but although looks natural, the answer is not factually correct (he won in 1976). Therefore, human confirmation of answer correctness demands additional effort and care due to the ability of LLMs to formulate these answers as complete and seemingly authoritative.

While it might be assumed that improved performance under lexical matching would reflect improved performance in an absolute sense, even if some correct answers are missed, we show this assumption does not hold. For this purpose, we manually re-evaluate several {\oqa} models on a random subset of {\NQopen} \cite{lee-etal-2019-latent}, an established benchmark. Not only is true performance substantially underestimated by this benchmark, but the relative performance of the models alters after re-evaluation: InstructGPT (zero-shot) achieves an accuracy of 12.6\% on our {\NQopen} subset, but our human judgment reveals its true performance to be 71.4\%, a nearly $+$60\% improvement.
Our linguistic analysis of the failure cases of lexical matching, an extension of a similar study by \citet{min2021neurips}, shows that the mismatches are mostly linguistically shallow and could be captured by simple patterns, such as regular expressions.

In contrast, automated evaluation mechanisms such as BEM~\cite{bulian-etal-2022-tomayto} based on semantic matching between the gold answers and generated answers produce a relative performance that is mostly consistent with human evaluation, although the absolute improvements are lower. However, long-form answers, generated by LLMs, introduce a new challenge that did not occur on prior models; they are prone to carry unattributable information \cite{rashkin2021measuring}. Automated evaluation models often deem the hallucinated responses correct, which is why, InstructGPT (zero-shot) is overestimated under these models, compared to human judgment.

We repeated this experiment with the 20-year-old CuratedTREC dataset \cite{trecqa2002} that provides its gold answers in the form of regular expressions. We observe that the relative performance of models remains mostly consistent under all three evaluation mechanisms, i.e., regular expressions, human evaluation, and semantic matching, with only slight differences in absolute performance.  However, the ranking discrepancy still persists between the two LLMs, i.e., InstructGPT (zero-shot) and InstructGPT (few-shot). 
Also, only under human judgment does the absolute performance of LLMs exceed that of the heavily engineered statistical NLP systems from 20 years ago on this collection.
Until recently, the best of these classical systems has been substantially superior to even the best of the modern neural models.
In light of our observations, we highlight that while semantic matching against exact answers would have been sufficient for QA evaluation prior to LLMs, they cannot accurately evaluate LLMs.

\section{Related Work}

\paragraph{Answer Equivalence in QA.} One way to tackle this task is through the automatic collection of alternative plausible answers from auxiliary knowledge sources such as a knowledge base \cite{si-etal-2021-whats}. However, the effectiveness of this approach is heavily contingent on the presence of answers in the knowledge source, which is often not the case. For instance, numerical answers or common phrases are unlikely to be found in a knowledge base. Moreover, matching gold answers with knowledge base entries can also be problematic as their surface forms may not be identical. Thus, these approaches fail to scale for various types of answers.
Another line of work focuses on building models to perform semantic similarity between candidate answers and gold answers, which can supersede lexical matching for verifying answers \cite{chen-etal-2019-evaluating, chen-etal-2020-mocha, risch-etal-2021-semantic,bulian-etal-2022-tomayto}. 
These methods indeed work well in reading comprehension because the presence of an input context often curtails the possibilities of models' generated answers. However, they are susceptible to failure in {\oqa} where questions should be answered without any additional context. Similarly, unsupervised semantic similarity-based evaluation metrics such as BERTScore~\cite{BERTScore} that rely on token-level matching of contextualized representations exhibit poor correlation with human judgment in QA evaluation \cite{chen-etal-2019-evaluating} and lack the ability to capture attributability \cite{maynez-etal-2020-faithfulness}.

\paragraph{Human Judgment in QA.}
Many works \cite{roberts-etal-2020-much,min2021neurips} resort to human evaluation to assess QA models. Although using humans for evaluation is expensive and not scalable, \citet{min2021neurips} find that the performance of QA systems bumps up 23\% on average using human judgment. The substantial gap between the true performance and token-based metrics showcases the long known strictness problem of lexical matching.

\section{{\Oqa} Evaluation}
The task of {\oqa} is referred to finding answers for information-seeking questions given a massive knowledge source such as Wikipedia \cite{voorhees-tice-2000-trec}. The questions are typically factoid with short answers and acontextual \cite{rogers2021qa}. {\Oqa} datasets encompass questions with their annotated gold answers that serve as a reference for evaluation.
Following reading comprehension \cite{rajpurkar-etal-2016-squad}, evaluation is carried out via lexical matching using the following two widely used metrics to measure the performance of models:

\begin{itemize}
\item \textbf{Exact-Match accuracy (EM)}: A candidate answer is deemed correct iff it can be found in the set of gold answers. The ratio of correct answers in the test collection is reported as EM accuracy.
\item \textbf{{\boldfone} score}: Considering answers as bags of tokens, a candidate answer receives a partial score ({\fone}) iff its tokens overlap with those of a gold answer. The maximum {\fone} score over a set of gold answers is assigned to the candidate answer. The final metric at corpus-level is measured via averaging {\fone} scores over the test collection.
\end{itemize}

\noindent
Based on the implementation of \citet{rajpurkar-etal-2016-squad}, answers are normalized (i.e., case-folded, and punctuation and articles are discarded) to compute these metrics.

\subsection{Models}
\label{sec:models}
We select {\oqa} models with publicly available codebase and reproduce their reported results. For all models, the ``base'' flavors are chosen for the experiments. In total, we use 12 models.

\paragraph{Retriever-Reader Models.} DPR~\cite{karpukhin-etal-2020-dense} is a well-known {\oqa} model that consists of a bi-encoder retriever and leverages an extractive reader. In addition to DPR, we pair several retrievers with Fusion-In-Decoder (FiD; \citealt{izacard-grave-2021-leveraging}), a prominent generative model that condition generating an answer on a list of passages: ANCE~\cite{ance}, Contriever\footnote{\url{https://huggingface.co/facebook/contriever-msmarco}}~\cite{contriever} RocketQAv2~\cite{ren-etal-2021-rocketqav2}, and FiD-KD~\cite{izacard2021distilling}. Further, we leverage GAR \cite{mao-etal-2021-generation}, a sparse retrieval model that augments questions with relevant contextual information generated by a fine-tuned T5 \cite{t5}. We fuse ANCE and GAR results with BM25, namely ANCE+ and GAR+, as they led to better results. We also use R2-D2 \cite{fajcik-etal-2021-r2-d2} that combines extractive and generative readers.

\paragraph{End-to-End Models.} EMDR$^2$~\cite{singh2021end} is an end-to-end model that jointly trains a dense retriever with a FiD-style reader. We also use EviGen~\cite{asai-etal-2022-evidentiality} that jointly learns to predict the evidentiality of passages and to generate the final answer in a multi-task fashion.

\paragraph{Closed-book Models.} We use InstructGPT\footnote{\texttt{text-davinci-003}, the details about this model are available at \url{https://beta.openai.com/docs/model-index-for-researchers}.} \cite{instructgpt} in two settings, following \citet{gpt3}: zero-shot and few-shot where the prompt includes 64 question/answer pairs, randomly sampled from the {\NQopen} training data.

\subsection{Dataset}
\label{sec:dataset}
We select questions from {\NQopen} \cite{lee-etal-2019-latent}, a popular {\oqa} benchmark, that consists of 3610 questions in the test set. We randomly sample 301 questions from {\NQopen}. Answers are generated via the prominent {\oqa} models, described in \S\ref{sec:models}, for the selected questions.
In total, the number of unique answers generated by the 12 models for 301 questions amounts to 1490 question/answer pairs.
Our experiments are done on Wikipedia, following the same settings provided by \citet{karpukhin-etal-2020-dense}.

\section{Strategies for Evaluating {\Oqa} Models}
Our goal is to shed light on the discrepancies between the actual and the measured accuracy of {\oqa} models. To this end, we adopt three evaluation mechanisms in addition to lexical matching to assess 12 {\oqa} models and draw a comparison between their estimated accuracy and the token-based performance.

\subsection{Supervised Evaluation via Semantic Similarity}
A common paradigm to evaluate QA systems is to cast evaluation as a classification task where the goal is to decide whether gold answers and candidate answers are semantically equivalent or not \cite{risch-etal-2021-semantic,bulian-etal-2022-tomayto}. To this end, we use a recent BERT-based model, namely BEM~\cite{bulian-etal-2022-tomayto}, that is trained on a human-annotated collection of answer pairs given a question, derived from SQuAD~\cite{rajpurkar-etal-2016-squad}. For evaluation, we feed a question along with a gold answer and a candidate answer to BEM and take its prediction. For questions with multiple gold answers, each gold answer is independently tested with a candidate answer. Once matched with either of the gold answers, a candidate answer is deemed correct.

\subsection{Zero-shot Evaluation via Prompting}
\label{sec:llm-eval}
We also test the ability of LLMs for evaluating QA models. In {\oqa}, the task of answer equivalence requires supplementary information in the absence of a given context, e.g., matching ``\textit{Jicheng}'' with ``\textit{Peking}'' in Figure~\ref{fig:examples}; therefore, LLMs are a reasonable choice here because they are equipped with an implicit memory that encompass knowledge \cite{roberts-etal-2020-much}, serving thus as an auxiliary information. To use LLMs for evaluating models, we elicit the following prompt through InstructGPT~\cite{instructgpt}:

\begin{quote}
    \small{
    \texttt{Question: what was the city of Beijing previously known as?}
    
    \texttt{Answer: Peking}
    
    \texttt{Candidate: Jicheng}

    \texttt{Is candidate correct?}
    }
\end{quote}

\noindent
We include the gold answer along with the candidate answer in the prompt, akin to the semantic similarity mechanism, as the objective here is to verify the correctness of the candidate. We call this evaluation method, InstructGPT-eval. We also test GPT-4~\cite{gpt4} using the same evaluation method, namely GPT4-eval, and observe that its results, reported in \S\ref{sec:gpt4}, closely resemble to those obtained from InstructGPT-eval.

\subsection{Human Evaluation}
\label{sec:human-eval} 
Human evaluation reflects the true performance of a model and serves as a basis for checking the feasibility of other evaluation mechanisms.
For this purpose, we ask two human annotators\footnote{The human annotators are the authors of this paper.} to judge whether a given answer to a question is correct or not. 
We present only question/answer pairs to human annotators to avoid any inadvertent biases, i.e., the annotators do not know which answers correspond to which model nor do they know if an answer is a gold answer. 
Annotators are allowed to use a search engine to find evidence that supports or rejects a candidate answer. Our annotation procedure is specifically geared towards {\oqa} unlike those of \citet{risch-etal-2021-semantic} and \citet{bulian-etal-2022-tomayto} that are designed for reading comprehension where annotators decide equivalence between a pair of answers given a question and a context.

The Fleiss' Kappa score between the two annotators is 72.8\%, i.e., 202 disagreements out of 1490 cases (13.6\%), indicating substantial agreement. 
Most disagreements arise from questions that are more likely to possess subjective answers. They mainly fall into three categories: ambiguous (e.g., ``{\em what is the corporate tax rate in great britain}''), list-style (e.g. ``{\em who dies in the lost city of z}''), and time-dependent (e.g. ``{\em latest series of keeping up with the kardashians}'') questions.
We ask a third annotator to judge the 202 cases where the two annotators diverged and take a majority vote to determine the correctness. 
The accepted answers by the annotators are then added to the set of gold answers for the selected questions. 
We compute the accuracy of the 12 models after amending the gold answers and compare it with the original accuracy that is computed via lexical matching.

\begin{table*}[t]
    \centering
    \small
    \resizebox{\textwidth}{!}{
    \begin{tabular}{l|c|c c| c c| c c | c c | c c}
    \toprule
         \multirow{2}{*}{{\bf Model}} & \multirow{2}{*}{{\bf K}} & \multicolumn{2}{c|}{{\em Entire Data} (3.6K)} & \multicolumn{2}{c|}{{\em Sampled} (301)} & \multicolumn{2}{c|}{{\em BEM}} & \multicolumn{2}{c|}{{\em InstructGPT-eval}} & \multicolumn{2}{c}{{\em Human}} \\
         & & \textbf{EM} & {\boldfone} & \textbf{EM} & {\boldfone} & \textbf{Acc} & $\mathbf{\Delta}$ & \textbf{Acc} & $\mathbf{\Delta}$ & \textbf{Acc} & $\mathbf{\Delta}$ \\
    \midrule
        InstructGPT (zero-shot) & - & 14.6 & - & 12.6 & 27.5 & 63.5 & \textbf{+50.9} & \textbf{77.1} & \textbf{+64.5} & 71.4 & \textbf{+58.8}  \\
        InstructGPT (few-shot) & - & 29.9 & - & 33.9 & 50.5 & 59.5 & +25.6 & 67.8 & +33.9 & \textbf{75.8} & +41.9 \\
        DPR & 50 & 40.9 & 47.8 & 45.9 & 52.3 & 52.5 & +6.6 & 55.1 & +9.2 & 58.8 & +12.9 \\
        FiD & 100 & 46.5 & 53.7 & 47.8 & 55.4 & 58.1 & +10.3 & 61.5 & +13.7 & 64.8 & +17.0 \\
        ANCE+ \& FiD & 50 & 47.3 & 54.8 & 48.2 & 55.9 & 59.5 & +11.3 & 63.1 & +14.9 & 65.8 & +17.6 \\
        RocketQAv2 \& FiD & 100 & 47.7 & 55.6 & 49.8 & 58.7 & 62.5 & +12.7 & 66.1 & +16.3 & 70.1 & +20.3 \\
        Contriever \& FiD & 100 & 47.9 & 55.4 & 46.5 & 55.9 & 60.8 & +14.3 & 63.1 & +16.6 & 66.5 & +20.0 \\
        FiD-KD & 100 & 49.6 & 57.4 & 50.8 & 61.2 & \textbf{65.8} & +15.0 & 70.4 & +19.6 & 73.1 & +22.3 \\
        GAR+ \& FiD & 100 & 49.8 & 57.4 & 50.8 & 59.7 & 63.1 & +12.3 & 67.1 & +16.3 & 69.4 & +18.2 \\
        EviGen & 20 & 49.8 & 57.0 & 51.8 & 59.5 & 62.1 & +10.3 & 64.8 & +13.0 & 67.1 & +15.3 \\
        EMDR$^2$ & 50 & 51.5 & \textbf{59.5} & \textbf{53.2} & \textbf{62.6} & 64.5 & +11.3 & 68.4 & +15.2 & 73.1 & +19.9 \\
        R2-D2 & 25 & \textbf{52.4} & 59.0 & 52.8 & 61.4 & 63.8 & +11.0 & 68.4 & +15.6 & 71.4 & +18.6 \\
    \bottomrule
    \end{tabular}}
    \caption{Accuracy of several {\oqa} models on a randomly sampled subset of 301 questions from {\NQopen} using lexical matching and the three evaluation mechanisms along with the reported results of these models on the entire dataset. 
    \textbf{K} refers to the number of passages fed to a model to generate an answer.
    InstructGPT (zero-shot) and InstructGPT (few-shot) achieve the highest raise in accuracy across all three additional evaluation methods. Only under human assessment does InstructGPT (few shot) outperform all other models.}
    \label{tab:NQOpen-eval}
\end{table*}

\begin{figure*}[t]
     \centering
     \begin{subfigure}[b]{0.32\textwidth}
        \centering
        \includegraphics[width=\textwidth]{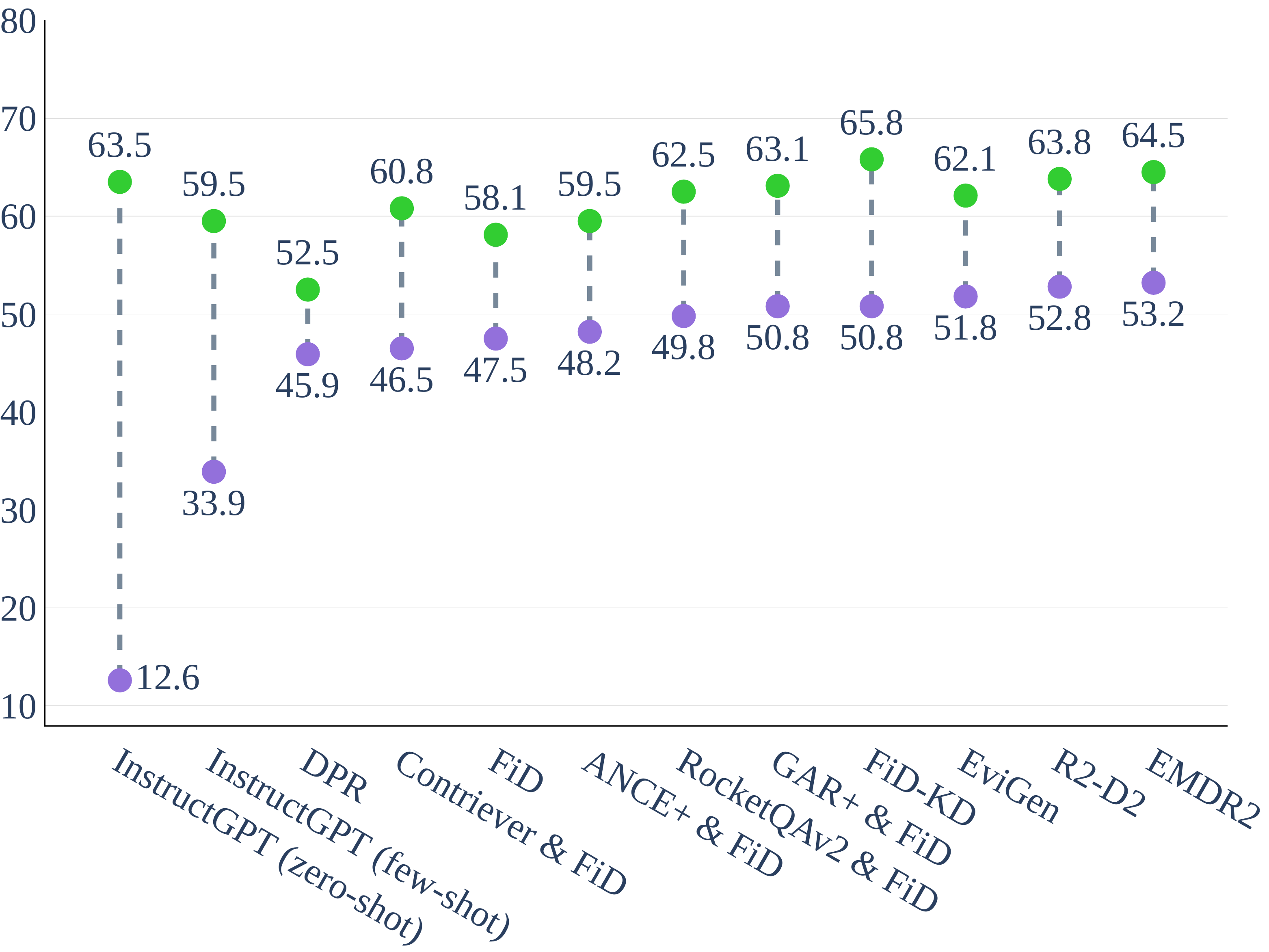}
        \caption{BEM}
        \label{fig:NQopen-dumbbell-bem}
    \end{subfigure}
    \hfill
    \begin{subfigure}[b]{0.32\textwidth}
        \centering
        \includegraphics[width=\textwidth]{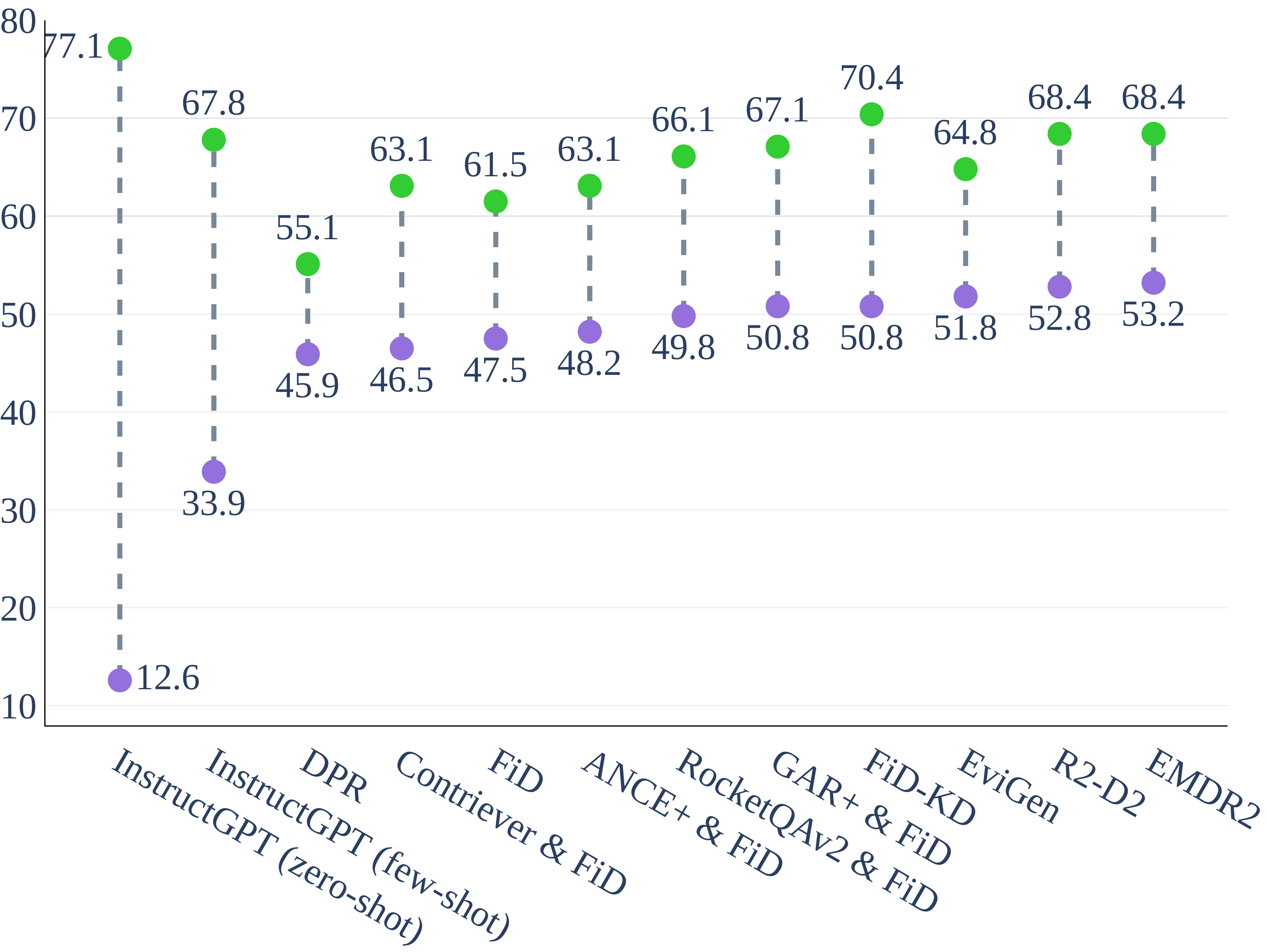}
        \caption{InstructGPT-eval}
        \label{fig:NQopen-dumbbell-gpt3}
    \end{subfigure}
    \hfill
    \begin{subfigure}[b]{0.32\textwidth}
         \centering
         \includegraphics[width=\textwidth]{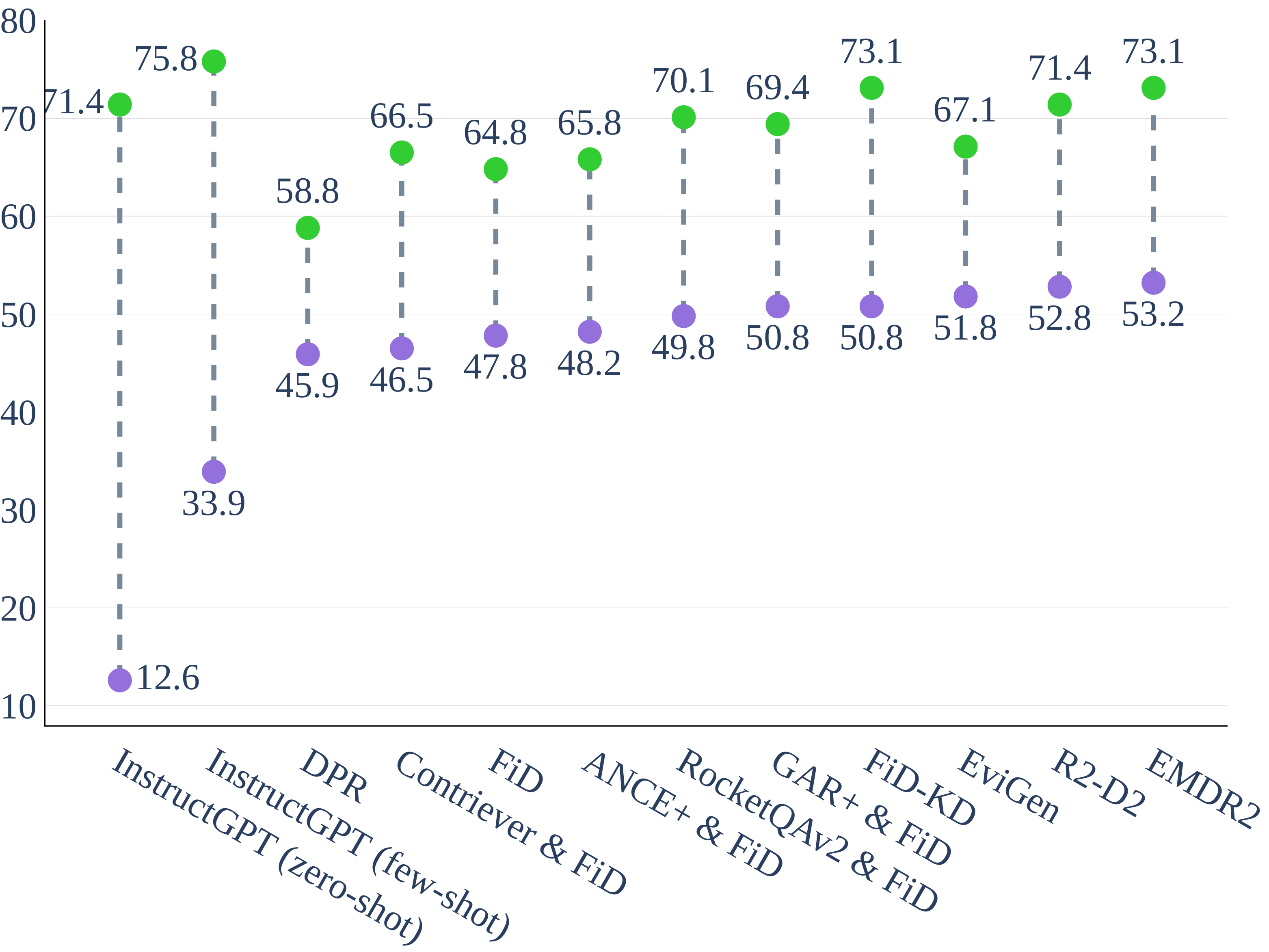}
         \caption{Human}
         \label{fig:NQopen-dumbbell-human}
    \end{subfigure}
    
    \caption{Accuracy of 12 {\oqa} models on the {\NQopen} subset of 301 questions using EM (purple points) and the three evaluation mechanisms (green points). For LLMs, the ranking of models under BEM and InstructGPT-eval is not consistent with human evaluation, while the rest of the models are ranked similarly under the two evaluation method. InstructGPT (few shot) outperforms other models only under human assessment.}
    \label{fig:NQopen-dumbbells}
\end{figure*}

\subsection{Results and Discussion}
Table~\ref{tab:NQOpen-eval} presents the accuracy of the {\oqa} models, computed using the three evaluation mechanisms, BEM, InstructGPT-eval, and Human, compared to the de facto EM accuracy.
The accuracy of all models consistently surges across all three evaluation mechanisms, i.e., 16\%, 21\%, and 24\% on average for BEM, InstructGPT-eval, and Human, respectively. InstructGPT (zero-shot) and InstructGPT (few-shot) are the top 2 models with the highest raise in accuracy across the evaluation mechanisms, whereas the amended result of DPR achieves the lowest increase.
Moreover, the accuracy reported using BEM and InstructGPT-eval are yet lower than that of human judgment, i.e., trailing 7.6\% and 2.9\% on average across all {\oqa} models, respectively.

More importantly, the ranking of models is readjusted by applying the three evaluation mechanisms. Figure~\ref{fig:NQopen-dumbbells} visualizes the accuracy of the {\oqa} models before (using only EM) and after our evaluation. EMDR$^2$, originally the best performing model, loses the top spot to InstructGPT (few-shot) by a nearly $+$3\% margin using human evaluation. BEM picks FiD-KD as the best model, whereas the LLM-based evaluation method estimates the highest accuracy for InstructGPT (zero-shot). Also, the Kendall's $\tau$ correlation of InstructGPT-eval, and BEM with human evaluation is 0.75, and 0.70, respectively, whereas EM and {\fone} show a significantly weaker correlation of 0.23 and 0.37. 

In contrast to human evaluation, BEM and InstructGPT-eval show that InstructGPT (zero-shot) has 4\%, and 9\% advantage, respectively, over InstructGPT (few-shot). To further investigate this phenomenon, we manually examine the InstructGPT (zero-shot) generated answers that are deemed incorrect by humans. We identify 47 unattributable answers out of 86 answers. The generated answers of InstructGPT (zero-shot) tend to be long statements that offer supplementary information, which raises the risk of containing hallucinated content. InstructGPT-eval accepts 30 of those answers ($\sim$10\% error over the 301 questions), whereas BEM incorrectly predicts 18 ($\sim$6\% error) answers as correct. Interestingly, GPT4-eval performs better and misidentifies only 9 cases ($\sim$3\% error). Yet, these results highlight that the automated methods are prone to misjudging hallucinated long answers, essentially rendering them unreliable against answers generated by LLMs.

\section{Linguistic Analysis of Correct Answers}
\label{sec:linguistic}
In this section, we aim to examine model answers that are not considered correct based on EM, but are in fact acceptable according to our assessment. \citet{min2021neurips} conducted a similar analysis on 50 questions for the participating models in the EfficientQA competition at NeurIPS 2020. In line with this work, we provide an in-depth analysis on a broader scale using more recent models to emphasize the drawbacks of widely used lexical-based evaluation metrics and semantic similarity methods.
We further dissect the categories presented by \citet{min2021neurips} into more detailed sub-categories.
Specifically, we group the 493 question/answer pairs that are deemed correct by humans while cannot be matched with gold answers into hierarchical categories as follows:\footnote{Long answers, generated by LLMs, are annotated based solely on the parts that candidate answers are mentioned.}

\paragraph{Semantic Equivalence:} Model predictions and gold answers convey the same meaning while not matching verbatim:
\begin{itemize}
    \item[\textbf{(i)}] \textbf{Multinominal entities}, e.g., ``\textit{Bhimrao Ramji Ambedkar}'' and ``\textit{B. R. Ambedkar}.''
    \item[\textbf{(ii)}] \textbf{Synonymous answers}, e.g., ``\textit{a virtual reality simulator}'' and ``\textit{a virtual reality world}.''
    \item[\textbf{(iii)}] \textbf{More elaborate answers}, e.g., ``\textit{Typically , no}'' and ``\textit{not required in all jurisdictions}.''
    \item[\textbf{(iv)}] \textbf{Exact-Match in explanatory answers}, e.g., ``\textit{1995}'' and ``\textit{Michael Jordan returned to the NBA in 1995.}''
    \item[\textbf{(v)}] \textbf{Bridging/Abridging}, e.g., ``\textit{citizens}'' vs. ``\textit{ordinary citizens}'' or ``\textit{in the Gospel of Luke}'' vs. ``\textit{Gospel of Luke}.''
    \item[\textbf{(vi)}] \textbf{Tokenization mismatches}, especially in the presence of punctuation marks, e.g., ``\textit{s-block}'' and ``\textit{s - block}.''
\end{itemize}

\paragraph{Symbolic Equivalence:} In case of numeric answers, gold answers and predicted ones can be symbolically identical either exactly or approximately while their surface text differs, e.g., ``\textit{about 3.99 degrees}'' vs. ``\textit{3.97 degrees}'' or the year ``\textit{1524}'' vs. ``\textit{the 16th century}.''

\paragraph{Intrinsic Ambiguity in Questions:} Ambiguous questions have several interpretations, each of which can lead to different answers. \citet{min-etal-2020-ambigqa} found that ambiguity is prevalent in {\NQopen}. Unlike other categories, mismatches that stem from ambiguity are not rooted in answers and instead, arise from questions themselves. For instance, ``\textit{when does the next episode of iZombie air?}'' presupposes a reference point in time that can only be clarified within a context. Thus, both ``\textit{May 07, 2018}'' and ``\textit{February 26, 2018}'' are correct, depending on when the question is asked.

\paragraph{Granularity Discrepancies:} Predicted answers may appear at different granularity levels than the gold answers. This case often arises for answers indicating spatial or temporal references. Indeed, under different presuppositions, some granularity levels are more preferable than others. Nonetheless, both predictions and gold answers are valid. We further categorize this discrepancy into:
\begin{itemize}
    \item[\textbf{(i)}] \textbf{Temporal granularity discrepancy}, e.g., ``\textit{when was the 50th star added to the united states flag?}'' can be answered by both ``\textit{1960}'' and ``\textit{July 4, 1960}.''
    \item[\textbf{(ii)}] \textbf{Spatial granularity discrepancy}, e.g., both ``\textit{Camping World Stadium}'' and ``\textit{Orlando, Florida}'' answer the question ``\textit{where is the citrus bowl held this year?}''
\end{itemize}

\paragraph{List-style Questions:} Actual answers to these kinds of questions encompass a set of plausible answers that is not fully specified in gold answers. For these questions, model answers are deemed correct if they are among at least one gold answer. We broke this group down into:
\begin{itemize}
    \item[\textbf{(i)}] \textbf{List questions}, e.g., gold answers to ``\textit{list of strict nature reserve in the Philippines}'' consist of six locations that is by no means comprehensive.
    \item[\textbf{(ii)}] \textbf{Open-ended questions} such as ``\textit{what is an example of a government monopoly in the United States?}'' where ``\textit{the United States Postal Service},'' not listed among gold answers, is a correct answer.
    \item[\textbf{(iii)}] \textbf{Compound questions} ask about multiple pieces of information in one question. They are a special case of multi-hop questions \cite{yang-etal-2018-hotpotqa}, e.g., ``\textit{when was the canadian pacific railway started and finished?}'' where the gold answer is ``\textit{between 1881 and 1885}'' vs. ``\textit{Started in 1881 and finished in 1885.}'' that is a correct answer.
\end{itemize}

\paragraph{Incorrect Gold Answers:} Models produce correct answers, but gold annotations are incorrect. Mismatches in this category are a byproduct of data quality issues. For example, the answer to ``\textit{what is the largest ethnic group in Mexico today?}'' is annotated ``\textit{K'iche'}'', whereas the correct answer is ``\textit{Mestizos}.''

\begin{figure}[t!]
  \centering
  \includegraphics[width=\linewidth]{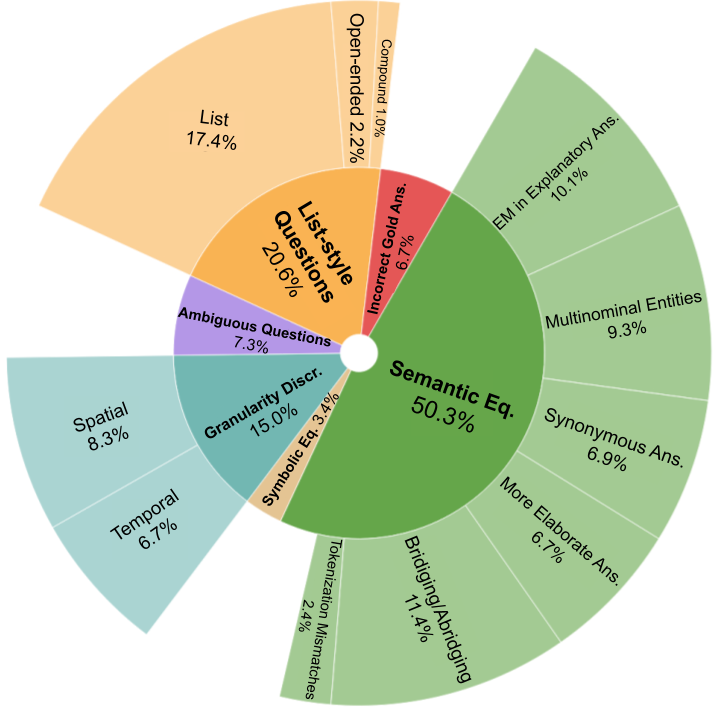}
  \caption{Statistics of exact-match failure modes determined via our linguistic analysis}
  \label{fig:linguistic}
\end{figure}

\subsection{Discussion}
\label{sec:discussion}
The statistics for each category are presented in Figure~\ref{fig:linguistic}. Semantic equivalence (50.3\%) is the most common failure mode of exact matching. The most frequent subcategories within this category are bridging/abridging (11.4\%), EM in explanatory answers (10.1\%), and multinominal entities (9.3\%). Other top frequent failure modes are list-style questions (20.6\%) and granularity discrepancy (15.0\%). Interestingly, most of these failure cases are related to syntactical variations of answers, which is why specifying gold answers via regular expressions can be useful in capturing these variations. Moreover, 14\% of EM failures are attributed to data quality issues, i.e., ambiguity and incorrect gold answers.

\begin{figure}[t]
  \centering
  \includegraphics[width=0.95\linewidth]{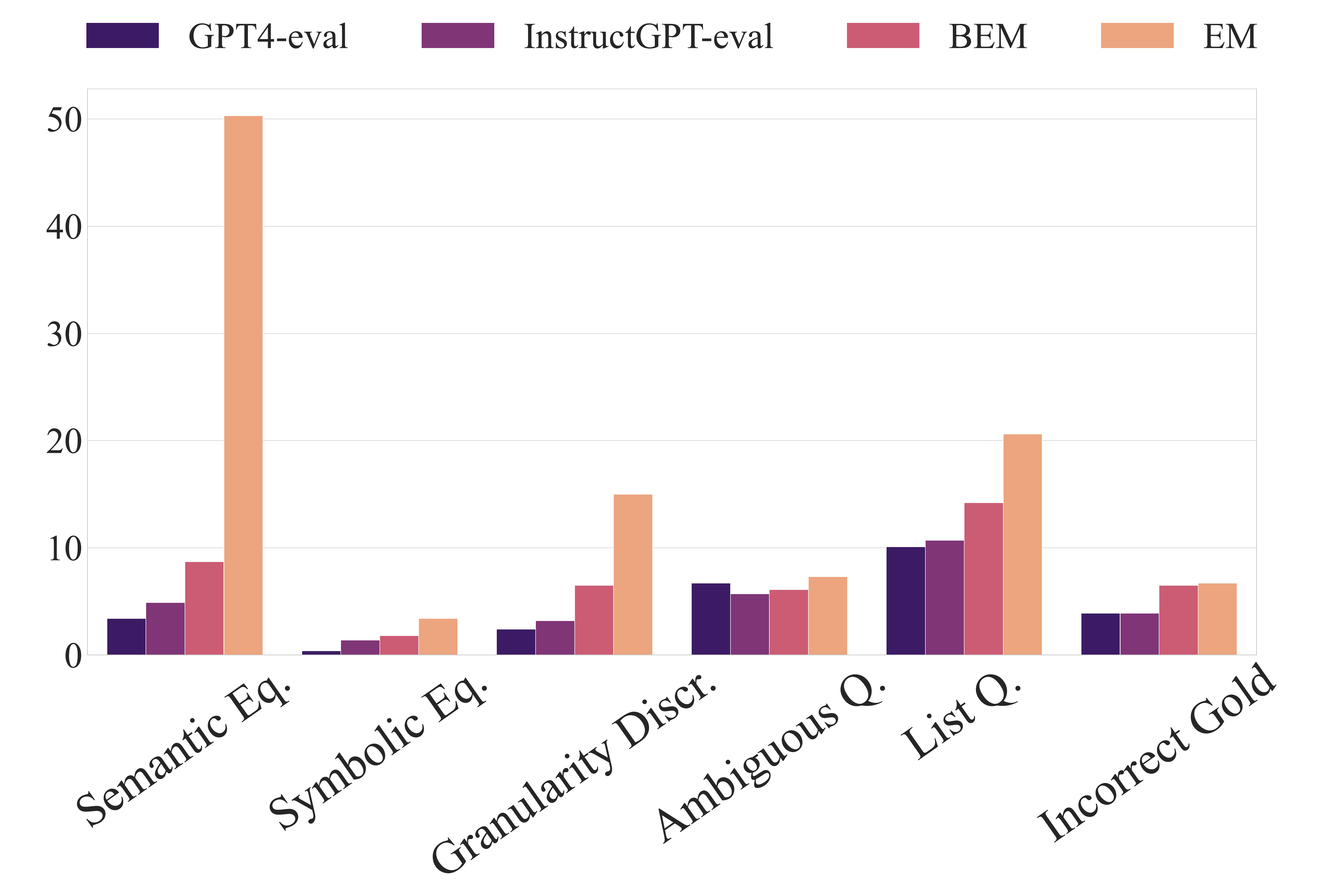}
  \caption{Percentage of high-level failure modes for each evaluation method on {\NQopen}.}
  \label{fig:NQopen-eval.analysis}
\end{figure}

\paragraph{Error Analysis of Automated Evaluation Methods.} The answers that InstructGPT-eval and BEM reject but humans consider correct are a subset of EM failures.\footnote{With only 3 exceptions: InstructGPT-eval rejects only 2 actually correct answers matching with gold answers that correspond to list questions where candidate answers appear in the middle of the gold answers. Moving the candidate answer to the top of the gold answer list would fix the issue. Similarly, BEM rejects only 1 exactly matched correct answer, i.e., ``{\em P-A-D-A-W-A-N.}'' while the gold answer is ``{\em Padawan}''.} More precisely, InstructGPT-eval and BEM reduce the 493 failure cases of EM to 149 (70\% $\downarrow$) and 217 (56\% $\downarrow$), respectively.
For GPT4-eval, the number of failure cases is 137 (72\%~$\downarrow$), only slightly lower than InstructGPT-eval.
The breakdown of the high-level failure categories for each evaluation method is shown in Figure~\ref{fig:NQopen-eval.analysis}. The three automated evaluation methods are able to fix most of the failures corresponding to semantic equivalence, granularity discrepancy, and symbolic equivalence. However, they do not perform that well on list-style questions where InstructGPT-eval and GPT4-eval still fail on more than 10\% of the EM failures, and BEM falls short on 14\%. They also perform nearly on par with EM on data quality-related failure cases, i.e., incorrect gold answers and ambiguous questions.

\begin{figure*}[t]
     \centering
     \begin{subfigure}[b]{0.32\textwidth}
        \centering
        \includegraphics[width=\textwidth]{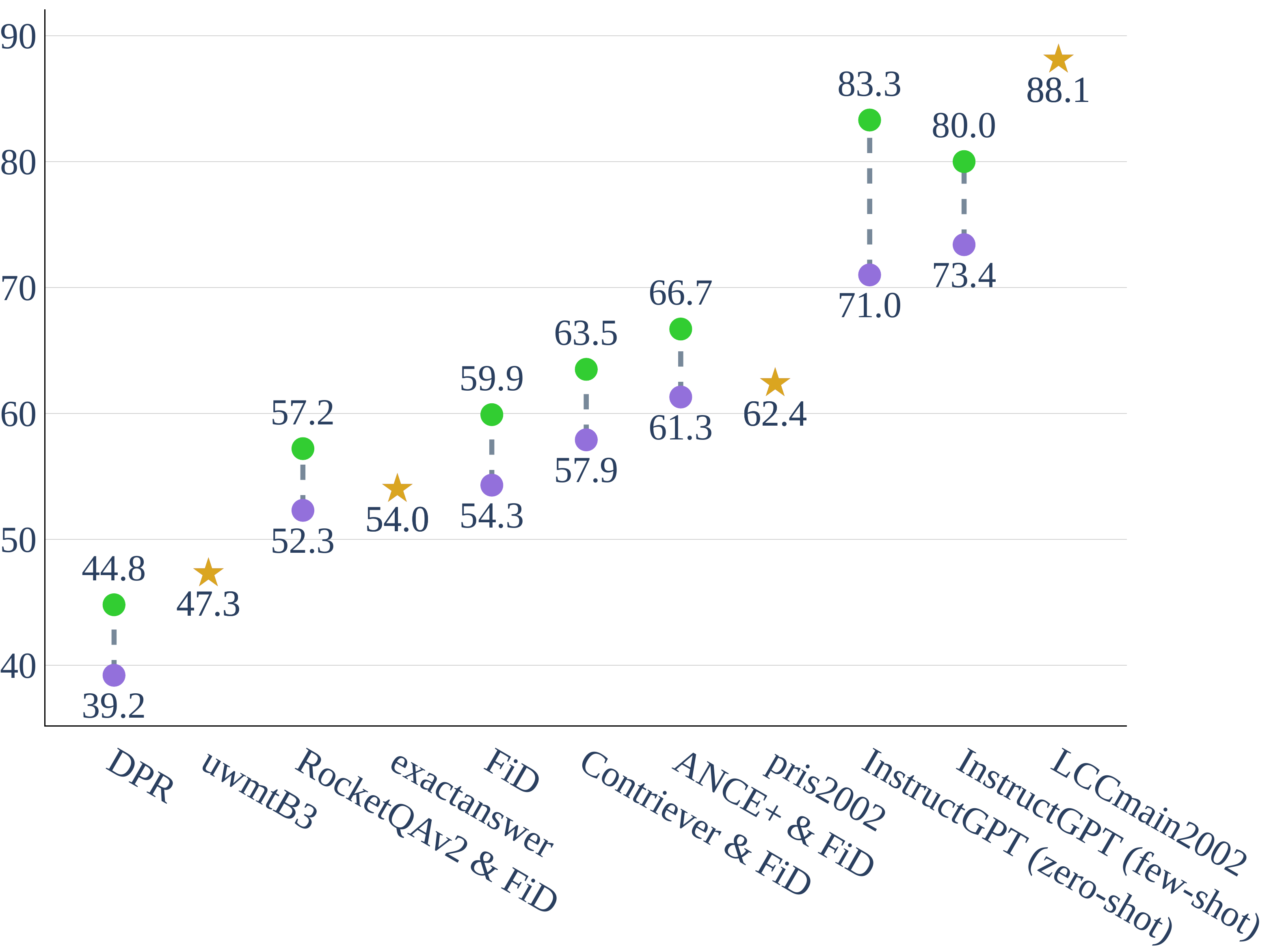}
        \caption{BEM}
        \label{fig:CuratedTREC-dumbbell-bem}
    \end{subfigure}
     \hfill
     \begin{subfigure}[b]{0.32\textwidth}
         \centering
         \includegraphics[width=\textwidth]{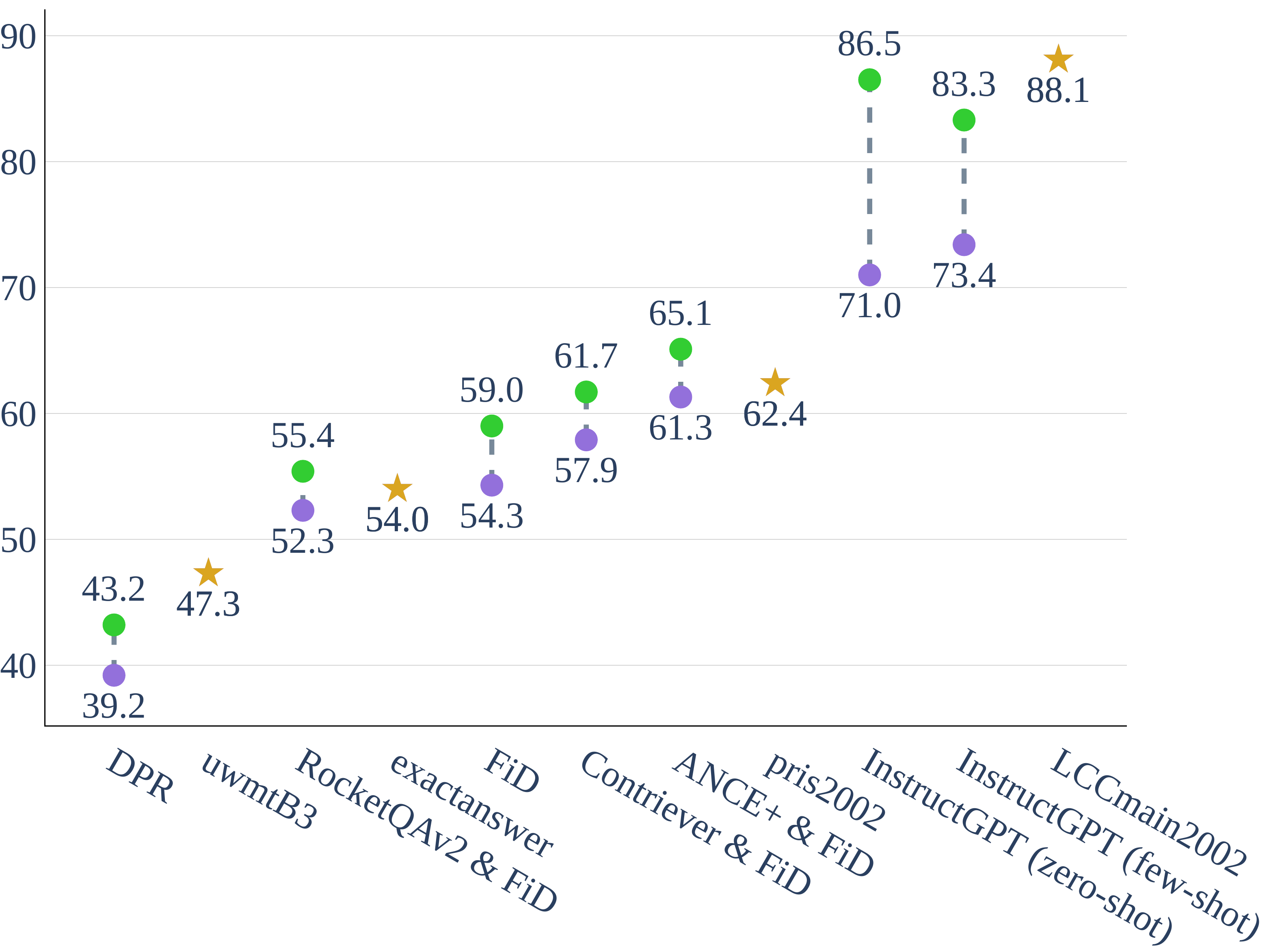}
         \caption{InstructGPT-eval}
         \label{fig:CuratedTREC-dumbbell-gpt3}
     \end{subfigure}
     \hfill
    \begin{subfigure}[b]{0.32\textwidth}
         \centering
         \includegraphics[width=\textwidth]{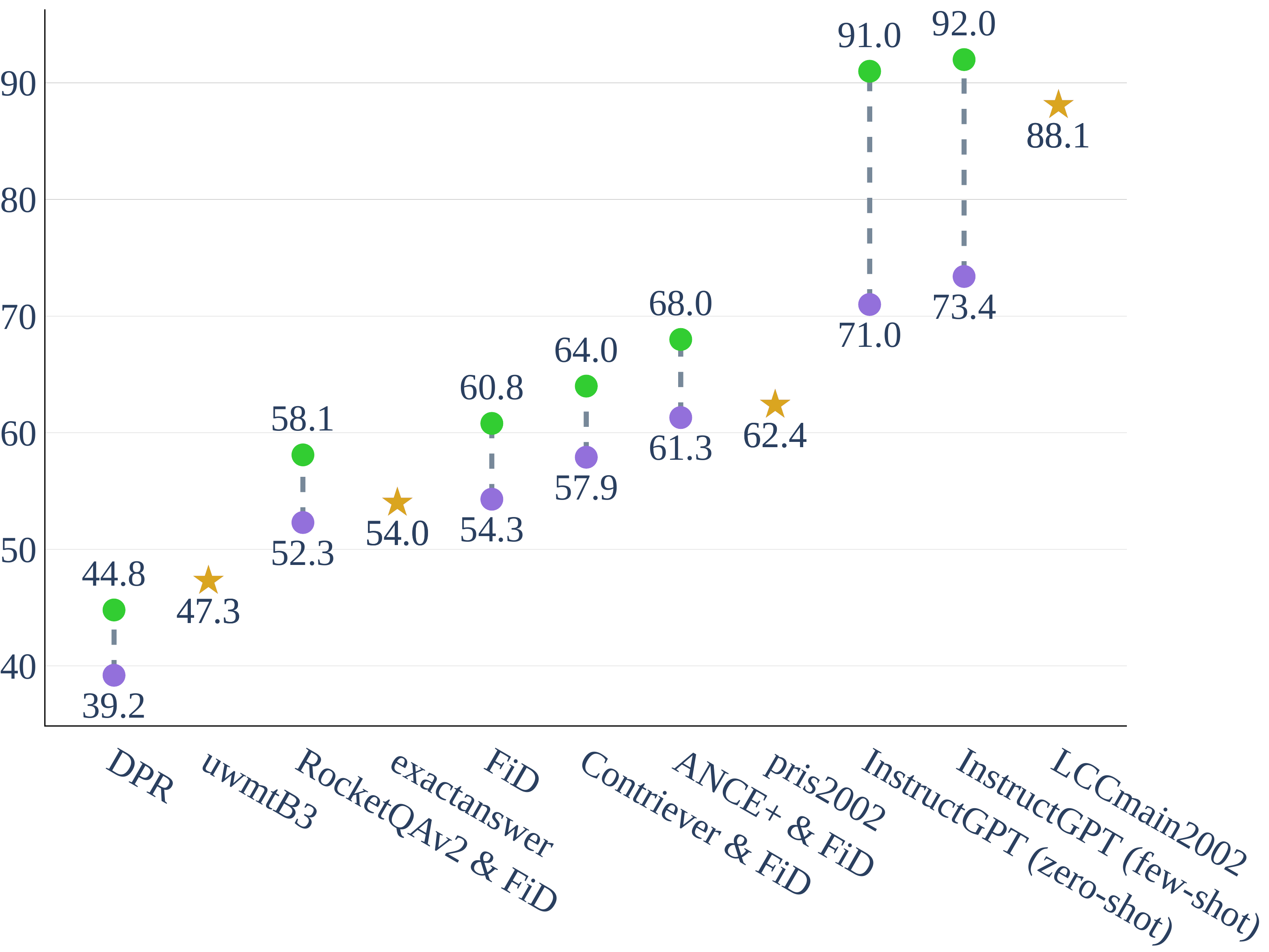}
         \caption{Human}
         \label{fig:CuratedTREC-dumbbell-human}
     \end{subfigure}
    
    \caption{Accuracy of several {\oqa} models on CuratedTREC 2002, computed via regex matching, along with the results of three evaluation mechanisms. Purple points represent the EM accuracy, and green points depict accuracy achieved via BEM, InstructGPT-eval, and human judgment. Classic statistical models from TREC QA 2002 are shown as orange stars. InstructGPT (few shot) outperforms the best of these classic models {\em only} under human assessment.}
    \label{fig:CuratedTREC-dumbbells}
\end{figure*}

\section{Regex Matching on CuratedTREC}

An alternative to lexical matching between gold answers and predicted answers during evaluation is to specify gold answers as regular expression patterns. Regex matching allows for capturing syntactical answer variations where exact-match falls short. In this section, our main goal is to highlight the advantages and pitfalls of using answer patterns in QA evaluation by comparing its results with our three evaluation mechanisms, described in \S\ref{sec:models}. 

\paragraph{Dataset.} We make a comparison across {\oqa} models on CuratedTREC 2002 \cite{baudivs2015modeling}, a dataset whose gold answers are specified by regular expressions. The questions in CuratedTREC are derived from the dataset in the QA tracks \cite{trecqa2002} of TREC 2001 to 2003 after a manual review to discard ambiguous or outdated questions. The knowledge source for TREC QA is originally English news text, namely AQUAINT, from three news sources (AP, NYTimes, and Xinhua), dating back to the late 90s.
Here, we opt for the original knowledge source to replicate the same environment as TREC QA 2002 so as to quantitatively gauge progress over two decade by comparing recent models with the models that took part in the QA track in 2002. This experiment is an out-of-distribution test for the neural models to check whether they are actually capable of using the knowledge source to answer questions or they answer from memory because the old news articles is less likely to have appeared in the pre-training corpus. However, LLMs inevitably do not use the knowledge source as they perform the task from their memory in a closed-book fashion. CuratedTREC 2002 consists of 444 questions whose answers are looked up in the AQAUINT corpus, comprising around 1M news articles. We follow \citet{karpukhin-etal-2020-dense} to split the articles into non-overlapping passages of 100 words, which amounts to over 4M passages in total. 

\paragraph{Models.} Out of the 12 models, we keep the ones that do not require further training on CuratedTREC 2002, leaving us with 7 models. These models produce 1872 unique answers on CuratedTREC 2002.
We also obtained the submitted run files of the participants in the TREC QA 2002 track from TREC organizers to compute their accuracy on CuratedTREC 2002. We include top 4 teams as baselines: LCCmain2002 (88.1\%; \citealt{pasca2001high}), pris2002 (62.4\%), exactanswer (54.0\%), and uwmtB3 (47.3\%).

Similar to {\NQopen}, we ask two annotators to judge 1872 question/answer pairs, followed by a third annotator who evaluates the diverging cases. The Fleiss' Kappa score between the two annotators is 83.5\%, i.e., 150 disagreements (8.0\%), indicating an almost perfect agreement.

The results are shown in Figure~\ref{fig:CuratedTREC-dumbbells}. Interestingly, the ranking of models via regex matching is left unchanged by all three evaluation mechanisms, except for InstructGPT (zero-shot) and InstructGPT (few-shot). Consistent with our observation on {\NQopen}, both BEM and InstructGPT-eval assign a higher accuracy to InstructGPT (zero-shot) over InstructGPT (few-shot). However, in contrast to {\NQopen}, they do not overestimate InstructGPT (zero-shot). Human evaluation shows that InstructGPT (few-shot), by scoring 92\%, is the best performing model, analogous to {\NQopen}. Among the non-LLM models, ANCE+ and Contriever consistently surpass other models. Similar to EM, regex matching is too rigid albeit to a lesser extent. In particular, the accuracy is underestimated by 6.6\%, 6.4\%, and 9.9\% on average via BEM, InstructGPT-eval, and human evaluation, respectively. 

We note that LCCmain2002, an original TREC run, outperforms all models prior to our assessment. Human evaluation highlights that both InstructGPT models are superior to LCCmain2002 by $+$1.9\% (for zero-shot) and $+$2.9\% (for few-shot). However, BEM and InstructGPT-eval fail to reflect this result. For other non-LLM models, ANCE+ and Contriever surpass pris2002 via all three evaluation methods (with the exception of Contriever using InstructGPT-eval). An interesting finding here is that although neural {\oqa} models are repeatedly proven to be powerful in accomplishing state-of-the-art, LCCmain2002, a heavily engineered statistical method from 20 years ago, ruffles their feathers by a substantial margin of 20\%. Only under human judgment does the absolute performance of LLMs surpass LCCmain2002.

\section{Conclusion}
Despite the simplicity and ubiquity of lexical matching as an evaluation method in {\oqa}, it is unnecessarily rigid because plausible candidate answers are likely not to appear in the list of gold answers. This flaw has been long known, but the efforts to circumvent it have been mostly artisanal. In this paper, we report a systematic study of lexical matching by manually judging answers generated by several prominent {\oqa} models. We found that LLMs achieve state-of-the-art on {\NQopen}. The accuracy of models is severely underestimated, 
with most EM failure cases stemming from syntactical variations of answers. 
Moreover, a zero-shot prompting method can be a reasonable substitute for human evaluation although it cannot detect unattributability in long-form answers.
Our insights and analysis in this paper will hopefully underpin the development of solid evaluation techniques in {\oqa}.

\section*{Limitations}
Our main focus in this work is limited to factoid information-seeking questions that typically prompt short answers. However, lexical matching is adopted by more complicated forms of QA that require complex reasoning. More precisely, QA tasks such as multi-hop reasoning \cite{yang-etal-2018-hotpotqa}, discrete reasoning \cite{dua-etal-2019-drop}, and causal relations \cite{lin-etal-2019-reasoning} also warrant similar systematic analysis as studied in this paper.

\section*{Acknowledgements}
We thank the anonymous reviewers for their constructive feedback.

\bibliography{anthology,main}

\begin{thebibliography}{45}
\expandafter\ifx\csname natexlab\endcsname\relax\def\natexlab#1{#1}\fi

\bibitem[{Asai et~al.(2022)Asai, Gardner, and
  Hajishirzi}]{asai-etal-2022-evidentiality}
Akari Asai, Matt Gardner, and Hannaneh Hajishirzi. 2022.
\newblock \href {https://doi.org/10.18653/v1/2022.naacl-main.162}
  {Evidentiality-guided generation for knowledge-intensive {NLP} tasks}.
\newblock In \emph{Proceedings of the 2022 Conference of the North American
  Chapter of the Association for Computational Linguistics: Human Language
  Technologies}, pages 2226--2243, Seattle, United States. Association for
  Computational Linguistics.

\bibitem[{Asai et~al.(2020)Asai, Hashimoto, Hajishirzi, Socher, and
  Xiong}]{asai2020Learning}
Akari Asai, Kazuma Hashimoto, Hannaneh Hajishirzi, Richard Socher, and Caiming
  Xiong. 2020.
\newblock \href {https://openreview.net/forum?id=SJgVHkrYDH} {Learning to
  retrieve reasoning paths over wikipedia graph for question answering}.
\newblock In \emph{International Conference on Learning Representations}.

\bibitem[{Baudi{\v{s}} and {\v{S}}ediv{\`y}(2015)}]{baudivs2015modeling}
Petr Baudi{\v{s}} and Jan {\v{S}}ediv{\`y}. 2015.
\newblock \href {https://doi.org/10.1007/978-3-319-24027-5_20} {Modeling of the
  question answering task in the {YodaQA} system}.
\newblock In \emph{International Conference of the cross-language evaluation
  Forum for European languages}, CLEF'15, pages 222--228. Springer-Verlag.

\bibitem[{Black et~al.(2022)Black, Biderman, Hallahan, Anthony, Gao, Golding,
  He, Leahy, McDonell, Phang, Pieler, Prashanth, Purohit, Reynolds, Tow, Wang,
  and Weinbach}]{black-etal-2022-gpt}
Sidney Black, Stella Biderman, Eric Hallahan, Quentin Anthony, Leo Gao,
  Laurence Golding, Horace He, Connor Leahy, Kyle McDonell, Jason Phang,
  Michael Pieler, Usvsn~Sai Prashanth, Shivanshu Purohit, Laria Reynolds,
  Jonathan Tow, Ben Wang, and Samuel Weinbach. 2022.
\newblock \href {https://doi.org/10.18653/v1/2022.bigscience-1.9}
  {{GPT}-{N}eo{X}-20{B}: An open-source autoregressive language model}.
\newblock In \emph{Proceedings of BigScience Episode {\#}5 -- Workshop on
  Challenges {\&} Perspectives in Creating Large Language Models}, pages
  95--136, virtual+Dublin. Association for Computational Linguistics.

\bibitem[{Brown et~al.(2020)Brown, Mann, Ryder, Subbiah, Kaplan, Dhariwal,
  Neelakantan, Shyam, Sastry, Askell, Agarwal, Herbert-Voss, Krueger, Henighan,
  Child, Ramesh, Ziegler, Wu, Winter, Hesse, Chen, Sigler, Litwin, Gray, Chess,
  Clark, Berner, McCandlish, Radford, Sutskever, and Amodei}]{gpt3}
Tom Brown, Benjamin Mann, Nick Ryder, Melanie Subbiah, Jared~D. Kaplan,
  Prafulla Dhariwal, Arvind Neelakantan, Pranav Shyam, Girish Sastry, Amanda
  Askell, Sandhini Agarwal, Ariel Herbert-Voss, Gretchen Krueger, Tom Henighan,
  Rewon Child, Aditya Ramesh, Daniel Ziegler, Jeffrey Wu, Clemens Winter, Chris
  Hesse, Mark Chen, Eric Sigler, Mateusz Litwin, Scott Gray, Benjamin Chess,
  Jack Clark, Christopher Berner, Sam McCandlish, Alec Radford, Ilya Sutskever,
  and Dario Amodei. 2020.
\newblock \href
  {https://proceedings.neurips.cc/paper/2020/file/1457c0d6bfcb4967418bfb8ac142f64a-Paper.pdf}
  {Language models are few-shot learners}.
\newblock In \emph{Advances in Neural Information Processing Systems},
  volume~33, pages 1877--1901. Curran Associates, Inc.

\bibitem[{Bulian et~al.(2022)Bulian, Buck, Gajewski, B{\"o}rschinger, and
  Schuster}]{bulian-etal-2022-tomayto}
Jannis Bulian, Christian Buck, Wojciech Gajewski, Benjamin B{\"o}rschinger, and
  Tal Schuster. 2022.
\newblock \href {https://aclanthology.org/2022.emnlp-main.20} {Tomayto,
  tomahto. beyond token-level answer equivalence for question answering
  evaluation}.
\newblock In \emph{Proceedings of the 2022 Conference on Empirical Methods in
  Natural Language Processing}, pages 291--305, Abu Dhabi, United Arab
  Emirates. Association for Computational Linguistics.

\bibitem[{Chen et~al.(2019)Chen, Stanovsky, Singh, and
  Gardner}]{chen-etal-2019-evaluating}
Anthony Chen, Gabriel Stanovsky, Sameer Singh, and Matt Gardner. 2019.
\newblock \href {https://doi.org/10.18653/v1/D19-5817} {Evaluating question
  answering evaluation}.
\newblock In \emph{Proceedings of the 2nd Workshop on Machine Reading for
  Question Answering}, pages 119--124, Hong Kong, China. Association for
  Computational Linguistics.

\bibitem[{Chen et~al.(2020)Chen, Stanovsky, Singh, and
  Gardner}]{chen-etal-2020-mocha}
Anthony Chen, Gabriel Stanovsky, Sameer Singh, and Matt Gardner. 2020.
\newblock \href {https://doi.org/10.18653/v1/2020.emnlp-main.528} {{MOCHA}: A
  dataset for training and evaluating generative reading comprehension
  metrics}.
\newblock In \emph{Proceedings of the 2020 Conference on Empirical Methods in
  Natural Language Processing (EMNLP)}, pages 6521--6532, Online. Association
  for Computational Linguistics.

\bibitem[{Chen et~al.(2017)Chen, Fisch, Weston, and
  Bordes}]{chen-etal-2017-reading}
Danqi Chen, Adam Fisch, Jason Weston, and Antoine Bordes. 2017.
\newblock \href {https://doi.org/10.18653/v1/P17-1171} {Reading {W}ikipedia to
  answer open-domain questions}.
\newblock In \emph{Proceedings of the 55th Annual Meeting of the Association
  for Computational Linguistics (Volume 1: Long Papers)}, pages 1870--1879,
  Vancouver, Canada. Association for Computational Linguistics.

\bibitem[{{Chowdhery} et~al.(2022){Chowdhery}, {Narang}, {Devlin}, {Bosma},
  {Mishra}, {Roberts}, {Barham}, {Chung}, {Sutton}, {Gehrmann}, {Schuh}, {Shi},
  {Tsvyashchenko}, {Maynez}, {Rao}, {Barnes}, {Tay}, {Shazeer}, {Prabhakaran},
  {Reif}, {Du}, {Hutchinson}, {Pope}, {Bradbury}, {Austin}, {Isard}, {Gur-Ari},
  {Yin}, {Duke}, {Levskaya}, {Ghemawat}, {Dev}, {Michalewski}, {Garcia},
  {Misra}, {Robinson}, {Fedus}, {Zhou}, {Ippolito}, {Luan}, {Lim}, {Zoph},
  {Spiridonov}, {Sepassi}, {Dohan}, {Agrawal}, {Omernick}, {Dai},
  {Sankaranarayana Pillai}, {Pellat}, {Lewkowycz}, {Moreira}, {Child},
  {Polozov}, {Lee}, {Zhou}, {Wang}, {Saeta}, {Diaz}, {Firat}, {Catasta}, {Wei},
  {Meier-Hellstern}, {Eck}, {Dean}, {Petrov}, and {Fiedel}}]{palm}
Aakanksha {Chowdhery}, Sharan {Narang}, Jacob {Devlin}, Maarten {Bosma}, Gaurav
  {Mishra}, Adam {Roberts}, Paul {Barham}, Hyung~Won {Chung}, Charles {Sutton},
  Sebastian {Gehrmann}, Parker {Schuh}, Kensen {Shi}, Sasha {Tsvyashchenko},
  Joshua {Maynez}, Abhishek {Rao}, Parker {Barnes}, Yi~{Tay}, Noam {Shazeer},
  Vinodkumar {Prabhakaran}, Emily {Reif}, Nan {Du}, Ben {Hutchinson}, Reiner
  {Pope}, James {Bradbury}, Jacob {Austin}, Michael {Isard}, Guy {Gur-Ari},
  Pengcheng {Yin}, Toju {Duke}, Anselm {Levskaya}, Sanjay {Ghemawat}, Sunipa
  {Dev}, Henryk {Michalewski}, Xavier {Garcia}, Vedant {Misra}, Kevin
  {Robinson}, Liam {Fedus}, Denny {Zhou}, Daphne {Ippolito}, David {Luan},
  Hyeontaek {Lim}, Barret {Zoph}, Alexander {Spiridonov}, Ryan {Sepassi}, David
  {Dohan}, Shivani {Agrawal}, Mark {Omernick}, Andrew~M. {Dai}, Thanumalayan
  {Sankaranarayana Pillai}, Marie {Pellat}, Aitor {Lewkowycz}, Erica {Moreira},
  Rewon {Child}, Oleksandr {Polozov}, Katherine {Lee}, Zongwei {Zhou}, Xuezhi
  {Wang}, Brennan {Saeta}, Mark {Diaz}, Orhan {Firat}, Michele {Catasta}, Jason
  {Wei}, Kathy {Meier-Hellstern}, Douglas {Eck}, Jeff {Dean}, Slav {Petrov},
  and Noah {Fiedel}. 2022.
\newblock \href {https://doi.org/10.48550/arXiv.2204.02311} {{PaLM}: Scaling
  language modeling with pathways}.
\newblock \emph{arXiv preprint arXiv:2204.02311}.

\bibitem[{Clark and Gardner(2018)}]{clark-gardner-2018-simple}
Christopher Clark and Matt Gardner. 2018.
\newblock \href {https://doi.org/10.18653/v1/P18-1078} {Simple and effective
  multi-paragraph reading comprehension}.
\newblock In \emph{Proceedings of the 56th Annual Meeting of the Association
  for Computational Linguistics (Volume 1: Long Papers)}, pages 845--855,
  Melbourne, Australia. Association for Computational Linguistics.

\bibitem[{Dua et~al.(2019)Dua, Wang, Dasigi, Stanovsky, Singh, and
  Gardner}]{dua-etal-2019-drop}
Dheeru Dua, Yizhong Wang, Pradeep Dasigi, Gabriel Stanovsky, Sameer Singh, and
  Matt Gardner. 2019.
\newblock \href {https://doi.org/10.18653/v1/N19-1246} {{DROP}: A reading
  comprehension benchmark requiring discrete reasoning over paragraphs}.
\newblock In \emph{Proceedings of the 2019 Conference of the North {A}merican
  Chapter of the Association for Computational Linguistics: Human Language
  Technologies, Volume 1 (Long and Short Papers)}, pages 2368--2378,
  Minneapolis, Minnesota. Association for Computational Linguistics.

\bibitem[{Dziri et~al.(2022)Dziri, Milton, Yu, Zaiane, and
  Reddy}]{dziri-etal-2022-origin}
Nouha Dziri, Sivan Milton, Mo~Yu, Osmar Zaiane, and Siva Reddy. 2022.
\newblock \href {https://doi.org/10.18653/v1/2022.naacl-main.387} {On the
  origin of hallucinations in conversational models: Is it the datasets or the
  models?}
\newblock In \emph{Proceedings of the 2022 Conference of the North American
  Chapter of the Association for Computational Linguistics: Human Language
  Technologies}, pages 5271--5285, Seattle, United States. Association for
  Computational Linguistics.

\bibitem[{Fajcik et~al.(2021)Fajcik, Docekal, Ondrej, and
  Smrz}]{fajcik-etal-2021-r2-d2}
Martin Fajcik, Martin Docekal, Karel Ondrej, and Pavel Smrz. 2021.
\newblock \href {https://doi.org/10.18653/v1/2021.findings-emnlp.73} {{R2-D2}:
  A modular baseline for open-domain question answering}.
\newblock In \emph{Findings of the Association for Computational Linguistics:
  EMNLP 2021}, pages 854--870, Punta Cana, Dominican Republic. Association for
  Computational Linguistics.

\bibitem[{Izacard et~al.(2022)Izacard, Caron, Hosseini, Riedel, Bojanowski,
  Joulin, and Grave}]{contriever}
Gautier Izacard, Mathilde Caron, Lucas Hosseini, Sebastian Riedel, Piotr
  Bojanowski, Armand Joulin, and Edouard Grave. 2022.
\newblock \href {https://openreview.net/forum?id=jKN1pXi7b0} {Unsupervised
  dense information retrieval with contrastive learning}.
\newblock \emph{Transactions on Machine Learning Research}.

\bibitem[{Izacard and Grave(2021{\natexlab{a}})}]{izacard2021distilling}
Gautier Izacard and Edouard Grave. 2021{\natexlab{a}}.
\newblock \href {https://openreview.net/forum?id=NTEz-6wysdb} {Distilling
  knowledge from reader to retriever for question answering}.
\newblock In \emph{International Conference on Learning Representations}.

\bibitem[{Izacard and
  Grave(2021{\natexlab{b}})}]{izacard-grave-2021-leveraging}
Gautier Izacard and Edouard Grave. 2021{\natexlab{b}}.
\newblock \href {https://doi.org/10.18653/v1/2021.eacl-main.74} {Leveraging
  passage retrieval with generative models for open domain question answering}.
\newblock In \emph{Proceedings of the 16th Conference of the European Chapter
  of the Association for Computational Linguistics: Main Volume}, pages
  874--880, Online. Association for Computational Linguistics.

\bibitem[{Karpukhin et~al.(2020)Karpukhin, Oguz, Min, Lewis, Wu, Edunov, Chen,
  and Yih}]{karpukhin-etal-2020-dense}
Vladimir Karpukhin, Barlas Oguz, Sewon Min, Patrick Lewis, Ledell Wu, Sergey
  Edunov, Danqi Chen, and Wen-tau Yih. 2020.
\newblock \href {https://doi.org/10.18653/v1/2020.emnlp-main.550} {Dense
  passage retrieval for open-domain question answering}.
\newblock In \emph{Proceedings of the 2020 Conference on Empirical Methods in
  Natural Language Processing (EMNLP)}, pages 6769--6781, Online. Association
  for Computational Linguistics.

\bibitem[{Khattab et~al.(2021)Khattab, Potts, and
  Zaharia}]{khattab-etal-2021-relevance}
Omar Khattab, Christopher Potts, and Matei Zaharia. 2021.
\newblock \href {https://doi.org/10.1162/tacl_a_00405} {Relevance-guided
  supervision for {O}pen{QA} with {C}ol{BERT}}.
\newblock \emph{Transactions of the Association for Computational Linguistics},
  9:929--944.

\bibitem[{Lee et~al.(2019)Lee, Chang, and Toutanova}]{lee-etal-2019-latent}
Kenton Lee, Ming-Wei Chang, and Kristina Toutanova. 2019.
\newblock \href {https://doi.org/10.18653/v1/P19-1612} {Latent retrieval for
  weakly supervised open domain question answering}.
\newblock In \emph{Proceedings of the 57th Annual Meeting of the Association
  for Computational Linguistics}, pages 6086--6096, Florence, Italy.
  Association for Computational Linguistics.

\bibitem[{Lin et~al.(2019)Lin, Tafjord, Clark, and
  Gardner}]{lin-etal-2019-reasoning}
Kevin Lin, Oyvind Tafjord, Peter Clark, and Matt Gardner. 2019.
\newblock \href {https://doi.org/10.18653/v1/D19-5808} {Reasoning over
  paragraph effects in situations}.
\newblock In \emph{Proceedings of the 2nd Workshop on Machine Reading for
  Question Answering}, pages 58--62, Hong Kong, China. Association for
  Computational Linguistics.

\bibitem[{Mao et~al.(2021)Mao, He, Liu, Shen, Gao, Han, and
  Chen}]{mao-etal-2021-generation}
Yuning Mao, Pengcheng He, Xiaodong Liu, Yelong Shen, Jianfeng Gao, Jiawei Han,
  and Weizhu Chen. 2021.
\newblock \href {https://doi.org/10.18653/v1/2021.acl-long.316}
  {Generation-augmented retrieval for open-domain question answering}.
\newblock In \emph{Proceedings of the 59th Annual Meeting of the Association
  for Computational Linguistics and the 11th International Joint Conference on
  Natural Language Processing (Volume 1: Long Papers)}, pages 4089--4100,
  Online. Association for Computational Linguistics.

\bibitem[{Maynez et~al.(2020)Maynez, Narayan, Bohnet, and
  McDonald}]{maynez-etal-2020-faithfulness}
Joshua Maynez, Shashi Narayan, Bernd Bohnet, and Ryan McDonald. 2020.
\newblock \href {https://doi.org/10.18653/v1/2020.acl-main.173} {On
  faithfulness and factuality in abstractive summarization}.
\newblock In \emph{Proceedings of the 58th Annual Meeting of the Association
  for Computational Linguistics}, pages 1906--1919, Online. Association for
  Computational Linguistics.

\bibitem[{Min et~al.(2021)Min, Boyd-Graber, Alberti, Chen, Choi, Collins, Guu,
  Hajishirzi, Lee, Palomaki, Raffel, Roberts, Kwiatkowski, Lewis, Wu,
  K\"uttler, Liu, Minervini, Stenetorp, Riedel, Yang, Seo, Izacard, Petroni,
  Hosseini, Cao, Grave, Yamada, Shimaoka, Suzuki, Miyawaki, Sato, Takahashi,
  Suzuki, Fajcik, Docekal, Ondrej, Smrz, Cheng, Shen, Liu, He, Chen, Gao, Oguz,
  Chen, Karpukhin, Peshterliev, Okhonko, Schlichtkrull, Gupta, Mehdad, and
  Yih}]{min2021neurips}
Sewon Min, Jordan Boyd-Graber, Chris Alberti, Danqi Chen, Eunsol Choi, Michael
  Collins, Kelvin Guu, Hannaneh Hajishirzi, Kenton Lee, Jennimaria Palomaki,
  Colin Raffel, Adam Roberts, Tom Kwiatkowski, Patrick Lewis, Yuxiang Wu,
  Heinrich K\"uttler, Linqing Liu, Pasquale Minervini, Pontus Stenetorp,
  Sebastian Riedel, Sohee Yang, Minjoon Seo, Gautier Izacard, Fabio Petroni,
  Lucas Hosseini, Nicola~De Cao, Edouard Grave, Ikuya Yamada, Sonse Shimaoka,
  Masatoshi Suzuki, Shumpei Miyawaki, Shun Sato, Ryo Takahashi, Jun Suzuki,
  Martin Fajcik, Martin Docekal, Karel Ondrej, Pavel Smrz, Hao Cheng, Yelong
  Shen, Xiaodong Liu, Pengcheng He, Weizhu Chen, Jianfeng Gao, Barlas Oguz,
  Xilun Chen, Vladimir Karpukhin, Stan Peshterliev, Dmytro Okhonko, Michael
  Schlichtkrull, Sonal Gupta, Yashar Mehdad, and Wen-tau Yih. 2021.
\newblock \href {https://proceedings.mlr.press/v133/min21a.html} {{NeurIPS 2020
  EfficientQA} competition: Systems, analyses and lessons learned}.
\newblock volume 133 of \emph{Proceedings of Machine Learning Research}, pages
  86--111. PMLR.

\bibitem[{Min et~al.(2020)Min, Michael, Hajishirzi, and
  Zettlemoyer}]{min-etal-2020-ambigqa}
Sewon Min, Julian Michael, Hannaneh Hajishirzi, and Luke Zettlemoyer. 2020.
\newblock \href {https://doi.org/10.18653/v1/2020.emnlp-main.466} {{A}mbig{QA}:
  Answering ambiguous open-domain questions}.
\newblock In \emph{Proceedings of the 2020 Conference on Empirical Methods in
  Natural Language Processing (EMNLP)}, pages 5783--5797, Online. Association
  for Computational Linguistics.

\bibitem[{{OpenAI}(2023)}]{gpt4}
{OpenAI}. 2023.
\newblock \href {https://doi.org/10.48550/arXiv.2303.08774} {{GPT-4} technical
  report}.
\newblock Technical report.

\bibitem[{Ouyang et~al.(2022)Ouyang, Wu, Jiang, Almeida, Wainwright, Mishkin,
  Zhang, Agarwal, Slama, Gray, Schulman, Hilton, Kelton, Miller, Simens,
  Askell, Welinder, Christiano, Leike, and Lowe}]{instructgpt}
Long Ouyang, Jeffrey Wu, Xu~Jiang, Diogo Almeida, Carroll Wainwright, Pamela
  Mishkin, Chong Zhang, Sandhini Agarwal, Katarina Slama, Alex Gray, John
  Schulman, Jacob Hilton, Fraser Kelton, Luke Miller, Maddie Simens, Amanda
  Askell, Peter Welinder, Paul Christiano, Jan Leike, and Ryan Lowe. 2022.
\newblock \href {https://openreview.net/forum?id=TG8KACxEON} {Training language
  models to follow instructions with human feedback}.
\newblock In \emph{Advances in Neural Information Processing Systems}, pages
  27730--27744. Curran Associates, Inc.

\bibitem[{Pasca and Harabagiu(2001)}]{pasca2001high}
Marius~A. Pasca and Sandra~M. Harabagiu. 2001.
\newblock \href {https://doi.org/10.1145/383952.384025} {High performance
  question/answering}.
\newblock In \emph{Proceedings of the 24th Annual International ACM SIGIR
  Conference on Research and Development in Information Retrieval}, SIGIR '01,
  page 366–374, New York, NY, USA. Association for Computing Machinery.

\bibitem[{Raffel et~al.(2020)Raffel, Shazeer, Roberts, Lee, Narang, Matena,
  Zhou, Li, and Liu}]{t5}
Colin Raffel, Noam Shazeer, Adam Roberts, Katherine Lee, Sharan Narang, Michael
  Matena, Yanqi Zhou, Wei Li, and Peter~J. Liu. 2020.
\newblock \href {http://jmlr.org/papers/v21/20-074.html} {Exploring the limits
  of transfer learning with a unified text-to-text transformer.}
\newblock \emph{Journal of Machine Learning Research}, 21(140):1--67.

\bibitem[{Rajpurkar et~al.(2016)Rajpurkar, Zhang, Lopyrev, and
  Liang}]{rajpurkar-etal-2016-squad}
Pranav Rajpurkar, Jian Zhang, Konstantin Lopyrev, and Percy Liang. 2016.
\newblock \href {https://doi.org/10.18653/v1/D16-1264} {{SQ}u{AD}: 100,000+
  questions for machine comprehension of text}.
\newblock In \emph{Proceedings of the 2016 Conference on Empirical Methods in
  Natural Language Processing}, pages 2383--2392, Austin, Texas. Association
  for Computational Linguistics.

\bibitem[{Rashkin et~al.(2021)Rashkin, Nikolaev, Lamm, Collins, Das, Petrov,
  Tomar, Turc, and Reitter}]{rashkin2021measuring}
Hannah Rashkin, Vitaly Nikolaev, Matthew Lamm, Michael Collins, Dipanjan Das,
  Slav Petrov, Gaurav~Singh Tomar, Iulia Turc, and David Reitter. 2021.
\newblock \href {https://doi.org/10.48550/arXiv.2112.12870} {Measuring
  attribution in natural language generation models}.
\newblock \emph{arXiv preprint arXiv:2112.12870}.

\bibitem[{Ren et~al.(2021)Ren, Qu, Liu, Zhao, She, Wu, Wang, and
  Wen}]{ren-etal-2021-rocketqav2}
Ruiyang Ren, Yingqi Qu, Jing Liu, Wayne~Xin Zhao, QiaoQiao She, Hua Wu, Haifeng
  Wang, and Ji-Rong Wen. 2021.
\newblock \href {https://doi.org/10.18653/v1/2021.emnlp-main.224}
  {{R}ocket{QA}v2: A joint training method for dense passage retrieval and
  passage re-ranking}.
\newblock In \emph{Proceedings of the 2021 Conference on Empirical Methods in
  Natural Language Processing}, pages 2825--2835, Online and Punta Cana,
  Dominican Republic. Association for Computational Linguistics.

\bibitem[{Risch et~al.(2021)Risch, M{\"o}ller, Gutsch, and
  Pietsch}]{risch-etal-2021-semantic}
Julian Risch, Timo M{\"o}ller, Julian Gutsch, and Malte Pietsch. 2021.
\newblock \href {https://doi.org/10.18653/v1/2021.mrqa-1.15} {Semantic answer
  similarity for evaluating question answering models}.
\newblock In \emph{Proceedings of the 3rd Workshop on Machine Reading for
  Question Answering}, pages 149--157, Punta Cana, Dominican Republic.
  Association for Computational Linguistics.

\bibitem[{Roberts et~al.(2020)Roberts, Raffel, and
  Shazeer}]{roberts-etal-2020-much}
Adam Roberts, Colin Raffel, and Noam Shazeer. 2020.
\newblock \href {https://doi.org/10.18653/v1/2020.emnlp-main.437} {How much
  knowledge can you pack into the parameters of a language model?}
\newblock In \emph{Proceedings of the 2020 Conference on Empirical Methods in
  Natural Language Processing (EMNLP)}, pages 5418--5426, Online. Association
  for Computational Linguistics.

\bibitem[{Rogers et~al.(2022)Rogers, Gardner, and Augenstein}]{rogers2021qa}
Anna Rogers, Matt Gardner, and Isabelle Augenstein. 2022.
\newblock \href {https://doi.org/10.1145/3560260} {{QA} dataset explosion: A
  taxonomy of {NLP} resources for question answering and reading
  comprehension}.
\newblock \emph{ACM Computing Surveys}, 55(10):1--45.

\bibitem[{Si et~al.(2021)Si, Zhao, and Boyd-Graber}]{si-etal-2021-whats}
Chenglei Si, Chen Zhao, and Jordan Boyd-Graber. 2021.
\newblock \href {https://doi.org/10.18653/v1/2021.emnlp-main.757} {What{'}s in
  a name? answer equivalence for open-domain question answering}.
\newblock In \emph{Proceedings of the 2021 Conference on Empirical Methods in
  Natural Language Processing}, pages 9623--9629, Online and Punta Cana,
  Dominican Republic. Association for Computational Linguistics.

\bibitem[{Singh et~al.(2021)Singh, Reddy, Hamilton, Dyer, and
  Yogatama}]{singh2021end}
Devendra Singh, Siva Reddy, Will Hamilton, Chris Dyer, and Dani Yogatama. 2021.
\newblock \href
  {https://proceedings.neurips.cc/paper/2021/file/da3fde159d754a2555eaa198d2d105b2-Paper.pdf}
  {End-to-end training of multi-document reader and retriever for open-domain
  question answering}.
\newblock In \emph{Advances in Neural Information Processing Systems},
  volume~34, pages 25968--25981.

\bibitem[{Voorhees(2003)}]{trecqa2002}
Ellen~M. Voorhees. 2003.
\newblock \href {https://tsapps.nist.gov/publication/get_pdf.cfm?pub_id=50780}
  {Overview of the {TREC} 2002 question answering track}.
\newblock In \emph{TREC}.

\bibitem[{Voorhees and Tice(2000)}]{voorhees-tice-2000-trec}
Ellen~M. Voorhees and Dawn~M. Tice. 2000.
\newblock \href {http://www.lrec-conf.org/proceedings/lrec2000/pdf/26.pdf} {The
  {TREC}-8 question answering track}.
\newblock In \emph{Proceedings of the Second International Conference on
  Language Resources and Evaluation ({LREC}{'}00)}, Athens, Greece. European
  Language Resources Association (ELRA).

\bibitem[{Wang et~al.(2018)Wang, Yu, Guo, Wang, Klinger, Zhang, Chang, Tesauro,
  Zhou, and Jiang}]{wang2018r}
Shuohang Wang, Mo~Yu, Xiaoxiao Guo, Zhiguo Wang, Tim Klinger, Wei Zhang, Shiyu
  Chang, Gerry Tesauro, Bowen Zhou, and Jing Jiang. 2018.
\newblock \href {https://doi.org/10.1609/aaai.v32i1.12053} {R\^{}3: Reinforced
  ranker-reader for open-domain question answering}.
\newblock In \emph{Proceedings of the AAAI Conference on Artificial
  Intelligence}.

\bibitem[{Xiong et~al.(2021)Xiong, Xiong, Li, Tang, Liu, Bennett, Ahmed, and
  Overwijk}]{ance}
Lee Xiong, Chenyan Xiong, Ye~Li, Kwok-Fung Tang, Jialin Liu, Paul~N. Bennett,
  Junaid Ahmed, and Arnold Overwijk. 2021.
\newblock \href {https://openreview.net/forum?id=zeFrfgyZln} {Approximate
  nearest neighbor negative contrastive learning for dense text retrieval}.
\newblock In \emph{International Conference on Learning Representations}.

\bibitem[{Yang et~al.(2018)Yang, Qi, Zhang, Bengio, Cohen, Salakhutdinov, and
  Manning}]{yang-etal-2018-hotpotqa}
Zhilin Yang, Peng Qi, Saizheng Zhang, Yoshua Bengio, William Cohen, Ruslan
  Salakhutdinov, and Christopher~D. Manning. 2018.
\newblock \href {https://doi.org/10.18653/v1/D18-1259} {{H}otpot{QA}: A dataset
  for diverse, explainable multi-hop question answering}.
\newblock In \emph{Proceedings of the 2018 Conference on Empirical Methods in
  Natural Language Processing}, pages 2369--2380, Brussels, Belgium.
  Association for Computational Linguistics.

\bibitem[{Ye and Durrett(2022)}]{ye2022the}
Xi~Ye and Greg Durrett. 2022.
\newblock \href
  {https://proceedings.neurips.cc/paper_files/paper/2022/file/c402501846f9fe03e2cac015b3f0e6b1-Paper-Conference.pdf}
  {The unreliability of explanations in few-shot prompting for textual
  reasoning}.
\newblock In \emph{Advances in Neural Information Processing Systems},
  volume~35, pages 30378--30392. Curran Associates, Inc.

\bibitem[{{Zhang} et~al.(2022){Zhang}, {Roller}, {Goyal}, {Artetxe}, {Chen},
  {Chen}, {Dewan}, {Diab}, {Li}, {Lin}, {Mihaylov}, {Ott}, {Shleifer},
  {Shuster}, {Simig}, {Singh Koura}, {Sridhar}, {Wang}, and
  {Zettlemoyer}}]{opt}
Susan {Zhang}, Stephen {Roller}, Naman {Goyal}, Mikel {Artetxe}, Moya {Chen},
  Shuohui {Chen}, Christopher {Dewan}, Mona {Diab}, Xian {Li}, Xi~Victoria
  {Lin}, Todor {Mihaylov}, Myle {Ott}, Sam {Shleifer}, Kurt {Shuster}, Daniel
  {Simig}, Punit {Singh Koura}, Anjali {Sridhar}, Tianlu {Wang}, and Luke
  {Zettlemoyer}. 2022.
\newblock \href {https://doi.org/10.48550/arXiv.2205.01068} {{OPT}: Open
  pre-trained transformer language models}.
\newblock \emph{arXiv preprint arXiv:2205.01068}.

\bibitem[{Zhang et~al.(2020)Zhang, Kishore, Wu, Weinberger, and
  Artzi}]{BERTScore}
Tianyi Zhang, Varsha Kishore, Felix Wu, Kilian~Q. Weinberger, and Yoav Artzi.
  2020.
\newblock \href {https://openreview.net/forum?id=SkeHuCVFDr} {{BERTScore}:
  Evaluating text generation with bert}.
\newblock In \emph{International Conference on Learning Representations}.

\end{thebibliography}
\bibliographystyle{acl_natbib}

\newpage
\clearpage
\appendix

\section{Zero-shot Evaluation using GPT-4}
\label{sec:gpt4}
For the sake of completeness, we test the ability of GPT-4 \cite{gpt4} for evaluating QA models as explained in \S\ref{sec:llm-eval}. We find that GPT4-eval results aligns with the trends observed in InstructGPT-eval, albeit displaying marginal improvements. Following the Table~\ref{tab:NQOpen-eval} layout, Table~\ref{tab:gpt4-NQOpen} presents the accuracy of the {\oqa} models, computed using GPT4-eval in conjunction with lexical matching, InstructGPT-eval, and human judgment as reference points.
The accuracy of all models consistently increases by an average of 20\% using GPT4-eval, which is similar to the increase level observed in InstructGPT-eval.
Moreover, analogous to InstructGPT-eval, the GPT4-eval accuracies are, on average, 3.3\% lower than those of human judgment.

\begin{figure}[ht]
 \centering
 \includegraphics[width=0.49\textwidth]{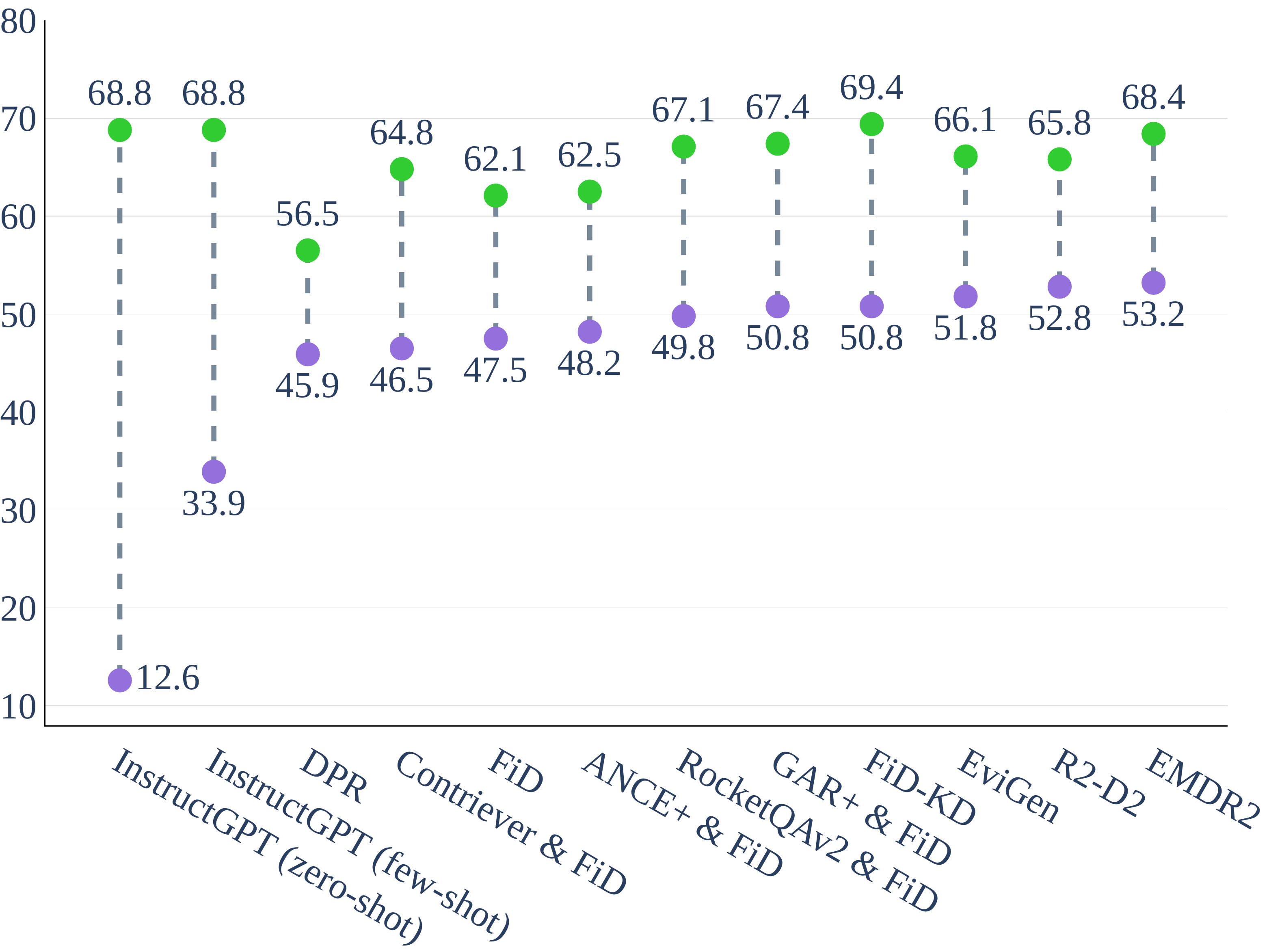}
 \caption{Accuracy of 12 {\oqa} models on the {\NQopen} subset of 301 questions using EM (purple points) and GPT4-eval (green points).}
 \label{fig:NQopen-dumbbell-gpt4}
\end{figure}

 Figure~\ref{fig:NQopen-dumbbell-gpt4} visualizes the accuracy of the {\oqa} models on {\NQopen} using EM and GPT4-eval, similar to Figure~\ref{fig:NQopen-dumbbells}. Unlike InstructGPT-eval, GPT4-eval estimates the highest accuracy for FiD-KD, followed by InstructGPT (zero-shot), InstructGPT (few-shot), and EMDR$^2$. Also, the Kendall's $\tau$ correlation of GPT4-eval with human judgment is 0.79, slightly higher than 0.75 of InstructGPT-eval. 

 \paragraph{Error Analysis:} As illustrated in Figure~\ref{fig:NQopen-eval.analysis}, GPT4-eval errors closely resemble the errors found in  InstructGPT-eval. However, for a small number of cases, GPT4-eval demonstrates unique erratic behaviours. First, for 2 cases, the model exhibits overconfidence in its internal memory and disregards gold answers that can be simply matched using EM. For example, GPT4-eval incorrectly rejects the candidate answer ``{\em Jermaine Jackson}'' (that is also a gold answer) to the question ``{\em Who sings Somebody's Watching Me with Michael Jackson?}'' We also observe the contradictory response of ``\textit{No, the candidate is correct}'' for 2 candidate answers that are correct, but are not included in the gold answers. Moreover, GPT4-eval incorrectly abstains from evaluating 2 candidate answers because it thinks more context is needed. For instance, it falsely utters 
 \begin{quote}
     ``{\em I cannot determine if the candidate is correct, as there is not enough information provided about the show "Fall" and the character Rose. Valene Kane is an actress, but without more context, it is unclear if she is related to this specific show or character.}''
 \end{quote}
 as a response to the question ``{\em Who is Rose in the Fall season 2?}'' and the candidate answer ``{\em Rose is a new character introduced in the second season of the show Fall. She is a mysterious woman who is connected to the supernatural events occurring in the town.}'' that is entirely fabricated.

\begin{figure}[ht]
  \centering
  \includegraphics[width=0.49\textwidth]{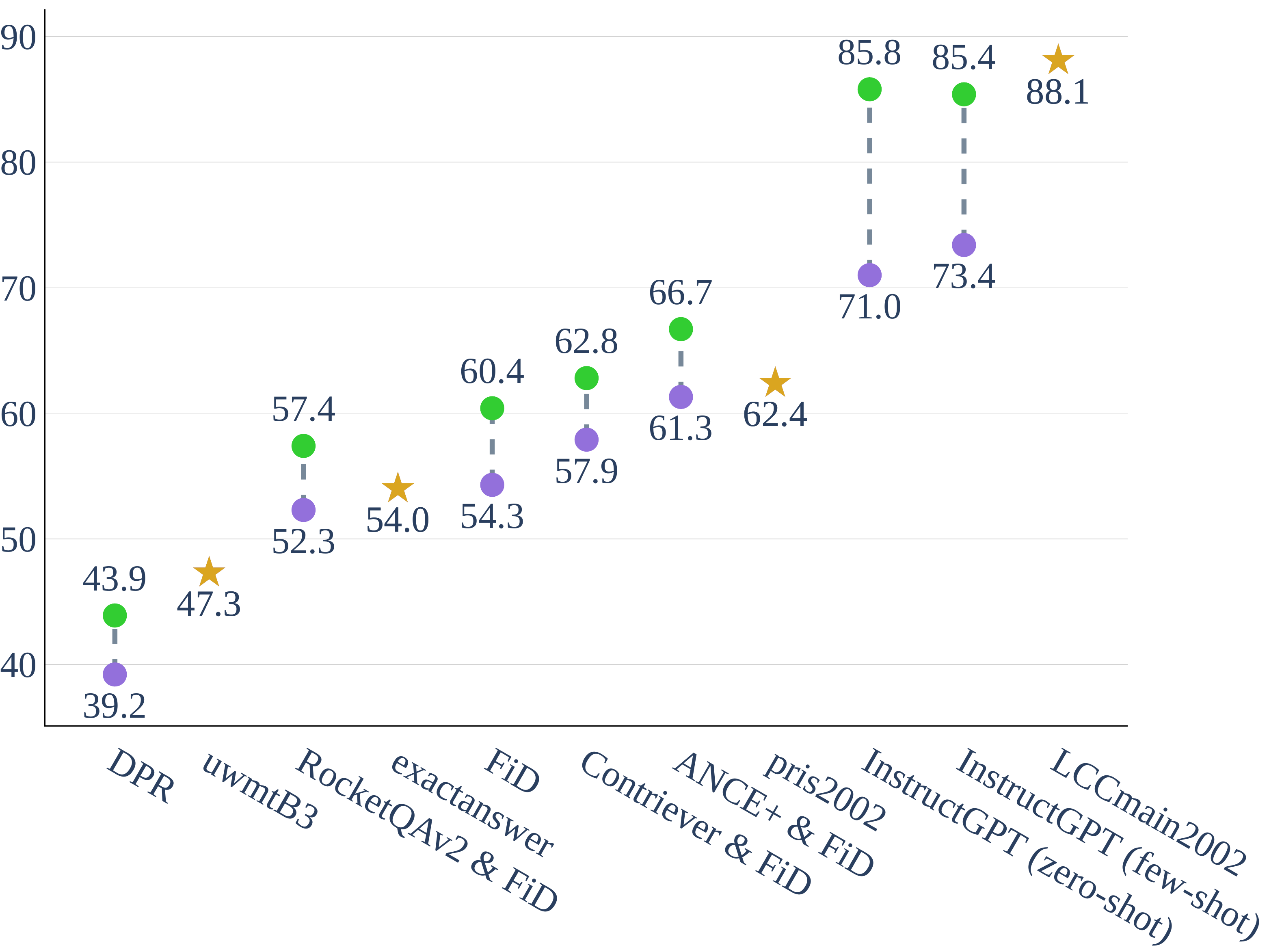}
  \caption{Accuracy of several {\oqa} models on CuratedTREC 2002, computed via regex matching (purple points), along with the results of GPT4-eval (green points), similar to Figure~\ref{fig:CuratedTREC-dumbbells}. Classic statistical models from TREC QA 2002 are shown as orange stars.}
  \label{fig:CuratedTREC-dumbbell-gpt4}
\end{figure}

\paragraph{Results on CuratedTREC 2002:} As shown in Figure~\ref{fig:CuratedTREC-dumbbell-gpt4}, GPT4-eval follows closely InstructGPT-eval on CuratedTREC 2002. Specifically, it indicates a higher accuracy for InstructGPT (zero-shot) compared to InstructGPT (few-shot) and ranks LCCmain2002 ahead of both InstructGPT models despite human evaluation suggesting otherwise.

\begin{table*}[ht]
    \resizebox{\textwidth}{!}{
    \begin{tabular}{l|c c| c c | c c | c c}
    \toprule
         \multirow{2}{*}{{\bf Model}} & \multicolumn{2}{c|}{{\em Sampled} (301)} & \multicolumn{2}{c|}{{\em InstructGPT-eval}} & \multicolumn{2}{c|}{{\em GPT4-eval}} & \multicolumn{2}{c}{{\em Human}} \\
         & \textbf{EM} & {\boldfone} & \textbf{Acc} & $\mathbf{\Delta}$ & \textbf{Acc} & $\mathbf{\Delta}$ & \textbf{Acc} & $\mathbf{\Delta}$ \\
    \midrule
        InstructGPT (zero-shot) & 12.6 & 27.5 & \textbf{77.1} & \textbf{+64.5} & 68.8 & \textbf{+56.2} & 71.4 & \textbf{+58.8}  \\
        InstructGPT (few-shot) & 33.9 & 50.5 & 67.8 & +33.9 & 68.8 & +34.9 & \textbf{75.8} & +41.9 \\
        DPR & 45.9 & 52.3 & 55.1 & +9.2 & 56.5 & +10.6 & 58.8 & +12.9 \\
        FiD & 47.8 & 55.5 & 61.5 & +13.7 & 61.8 & +14.0 & 64.8 & +17.0 \\
        ANCE+ \& FiD & 48.2 & 55.9 & 63.1 & +14.9 & 62.5 & +14.3 & 65.8 & +17.6 \\
        RocketQAv2 \& FiD & 49.8 & 58.7 & 66.1 & +16.3 & 67.1 & +17.3 & 70.1 & +20.3 \\
        Contriever \& FiD & 46.5 & 55.9 & 63.1 & +16.6 & 64.8& +18.3& 66.5 & +20.0 \\
        FiD-KD & 51.2 & 61.6 & 70.4 & +19.6 & \textbf{69.4} & +18.6 & 73.1 & +22.3 \\
        GAR+ \& FiD & 50.8 & 59.7 & 67.1 & +16.3 & 67.4 & +16.6 & 69.4 & +18.2 \\
        EviGen & 51.8 & 59.5 & 64.8 & +13.0 & 66.1 & +14.3 & 67.1 & +15.3 \\
        EMDR$^2$ & \textbf{53.2} & \textbf{62.6} & 68.4 & +15.2 & 68.4 & +15.2 & 73.1 & +19.9 \\
        R2-D2 & 52.8 & 61.4 & 68.4 & +15.6 & 65.8 & +13.0 & 71.4 & +18.6 \\
    \bottomrule
    \end{tabular}}
    \caption{Accuracy of several {\oqa} models on a randomly sampled subset of 301 questions from {\NQopen} using lexical matching, GPT4-eval, human evaluation. Only GPT4-eval results are new here. The rest of the results are already reported in Table~\ref{tab:NQOpen-eval} and copied here solely as a reference. GPT4-eval demonstrates approximately similar behaviour as InstructGPT-eval when ranking the models.}
    \label{tab:gpt4-NQOpen}
\end{table*}

\end{document}